\RequirePackage{rotating}

\documentclass[acmsmall,screen,table,usenames,dvipsnames]{acmart}

\AtBeginDocument{%
  \providecommand\BibTeX{{%
    \normalfont B\kern-0.5em{\scshape i\kern-0.25em b}\kern-0.8em\TeX}}}

\acmJournal{CSUR}
\acmVolume{0}
\acmNumber{0}
\acmArticle{0}
\acmMonth{0}

\usepackage[inline]{enumitem}
\usepackage{graphicx}

\newcommand{\eg}{e.g., }
\newcommand{\ie}{i.e., }

\newcommand{\figref}[1]{Fig.~\ref{#1}}    
\newcommand{\tabref}[1]{Table~\ref{#1}}
\newcommand{\Tabref}[1]{Table~\ref{#1}}
\newcommand{\secref}[1]{Section~\ref{#1}}

\usepackage{stmaryrd}
\usepackage{framed}

\usepackage{graphicx}
\usepackage{adjustbox}
\usepackage{listings}
\usepackage{caption,setspace}
\usepackage{diagbox} 
\usepackage{booktabs} 
\usepackage{rotating}
\usepackage{array,multirow}
\usepackage{hhline}
\usepackage{todonotes}
\usepackage{flushend}
\usepackage[inline]{enumitem}
\usepackage{csquotes}
\usepackage{nameref}
\usepackage{pifont}

\newcommand{\cmark}{\ding{51}}%
\newcommand{\xmark}{\ding{55}}%
\newcommand{\Fone}{{F$_{1}$}}

\newcommand{\revisedgiannis}[1]{\textcolor{black}{#1}}
\newcommand{\giannisrtwo}[1]{\textcolor{black}{#1}}
\newcommand{\giannisrthree}[1]{\textcolor{black}{#1}}

\newcommand{\nikos}[1]{\textcolor{black}{#1}}

\begin{document}

\title{A Review on Fact Extraction and Verification}

\author{Giannis Bekoulis}
\email{gbekouli@etrovub.be}
\affiliation{%
  \institution{ETRO \nikos{Department}, Vrije Universiteit Brussel}
  \streetaddress{Pleinlaan 2}
  \city{Brussels}
  \postcode{1050}
  \country{Belgium}
}
\affiliation{%
  \institution{imec}
  \streetaddress{Kapeldreef 75}
  \city{Leuven}  
  \postcode{3001}
  \country{Belgium}
}

\author{Christina Papagiannopoulou}
\email{cppapagi@gmail.com}

\author{Nikos Deligiannis}
\email{ndeligia@etrovub.be}
\affiliation{%
  \institution{ETRO \nikos{Department}, Vrije Universiteit Brussel}
  \streetaddress{Pleinlaan 2}
  \city{Brussels}  
  \postcode{1050}
  \country{Belgium}
}
\affiliation{%
  \institution{imec}
  \streetaddress{Kapeldreef 75}
  \city{Leuven}  
  \postcode{3001}
  \country{Belgium}
}



\begin{abstract}
\revisedgiannis{We study the fact checking problem, which aims to identify the veracity of a given claim. Specifically, we focus on the task of Fact Extraction and VERification (FEVER) and its accompanied dataset.} The task consists of the subtasks of retrieving the relevant documents (and sentences) from Wikipedia and validating whether the information in the documents supports or refutes a given claim. This task is essential and can be the building block of applications such as fake news detection and medical claim verification. In this paper, we aim at a better understanding of the challenges of the task by presenting the literature in a structured and comprehensive way. We describe the proposed methods by analyzing the technical perspectives of the different approaches and discussing the performance results on the FEVER dataset, which is the most well-studied and formally structured dataset on the fact extraction and verification task. We also conduct the largest experimental study to date on identifying beneficial loss functions for the sentence retrieval component. Our analysis indicates that sampling negative sentences is important for improving the performance and decreasing the computational complexity. Finally, we describe open issues and future challenges, and we motivate future research in the task.

\end{abstract}




\maketitle

\section{Introduction} 
Nowadays we  are confronted with a large amount of information of questionable origin or validity. This is not a new problem as it has \nikos{appeared} since the very
first years of the printing press. However, it attracted a growing interest with the wide use of social media streams as online news sources. On a daily basis, a large audience accesses various media outlets such as news blogs, etc., rapidly consuming a vast amount of information with possibly inaccurate or even misleading content. The proliferation of the misleading content happens really quickly due to the fast dissemination of news across various media streams.

Recently, a lot of research in the \nikos{natural language processing (NLP)} community has been focused on detecting whether the information coming from news sources is fake or not. \revisedgiannis{Specifically, automated fact checking is the NLP task that aims at determining the veracity of a given claim~\cite{vlachos-riedel-2014-fact}.} \giannisrthree{Since the units of a document are the sentences, a way to achieve the fact checking objective is to automatically identify the nature of the relationship between any two sentences} (\eg
if they contradict or support one another).  
To this end, such systems require a certain level of understanding the language and the coherence of textual units. Thus, the methods that have been proposed to solve the fact verification problem essentially belong to the broader family of Natural Language Inference (NLI) systems~\cite{bowman:15}. However, we should note that an NLI system (\ie used to validate the veracity of \nikos{a} claim) requires all the necessary information (\ie the claim and the potential evidence sentences which contradict or support the claim) to be available upfront. For this \nikos{reason}, retrieval systems are also important in order to identify the relevant and trustworthy evidence sentences coming from various sources (\eg Wikipedia).

\revisedgiannis{In this paper, we study the fact checking problem~\cite{vlachos-riedel-2014-fact}. In particular, we focus on the recently introduced task of FEVER and its accompanied dataset~\cite{thorne:18}.} Given an artificially constructed claim and Wikipedia documents, the goal of the task is to validate the veracity of the claim. For that, a dataset comprising 145,449 training claims has been introduced in the work of~\citet{thorne:18} and \revisedgiannis{two competitions (\ie shared tasks) have been organised~\cite{thorne:18b,thorne:19b}}, where several models have been proposed. Since then, the task has received a lot of attention from the NLP community and several complex architectures -- relying, for example, on Graph Neural Networks (GNNs)~\cite{kipf:17} \nikos{or} language models~\cite{delvin:19} -- have been presented in top-tier NLP venues to resolve the task (i.e., \giannisrthree{achieving scores as high as 86.74 points in terms of the FEVER score}~\cite{petroni2020kilt}, see~\secref{sec:evaluation:claimm_verification} for the definition of the FEVER score). 

\giannisrthree{This work aims at summarizing the methods introduced to resolve the FEVER task, which has the specific characteristic that it comprises multiple subtasks. 
Our goal is to present a comprehensive study and analysis of the methods that are used in the context of claim extraction and verification. 
We focus on the FEVER dataset due to that it is the most popular, well-studied and formally structured dataset about the task of claim extraction and verification.
There is a large number of survey papers~\cite{zhou:20review,SAQUETE2020112943,Sharma:19,BONDIELLI201938,oshikawa:20} for the related task of fake news detection, where its goal is to identify false information in the media~\cite{BONDIELLI201938}.
The fake news detection task is strongly related to automated fact checking and some of the previous review studies on fake news detection~\cite{oshikawa:20,BONDIELLI201938}) consider automated fact checking as a constituent of fake news detection.
Our work is mostly related to the work of~\citet{thorne:18c} and our survey paper focuses on the fact checking problem (and not in the more general problem of fake news detection).
Unlike the work of~\citet{thorne:18c}, which provides a high level overview of fact checking in terms of terminology and methods, we aim at thoroughly examining the methods that have been developed for the FEVER task from a more technical perspective. Specifically, we describe the various technologies used in the methods, highlight the pros and cons of each architecture, discuss the contribution of each specific component in each proposed system and present the results for each individual subtask of the FEVER task. That way,
we are able to compare each and every method by quantifying the benefit of each and every component of these methods. By comparing several methods on the same dataset, we are able to perform
a fair and direct comparison of the various architectures. Moreover, as mentioned above, the task consists of three subtasks, making it challenging enough to act as a common ground to evaluate and benchmark various approaches. In the various studies that try to solve the FEVER task, researchers use different evaluation metrics, data splits, assumptions and thus, it is not straightforward for one to compare them and identify similar and different aspects among them. This is exactly the gap that our work covers; namely, to present in a structured and comprehensive way the various methods introduced for the FEVER task. }

Our study serves as an extensive review for the
FEVER task and has as a goal to help future researchers to improve the current state-of-the-art by \nikos{1)} providing an easy way to compare systems and experiments, and \nikos{2)} investigating the contributions of each individual model component in the overall performance. 
\giannisrtwo{We also present related datasets, competitions, and recent work on explainability on the task of fact extraction and verification.}
On top of that -- to the best of our knowledge -- we conduct the largest experimental study on the sentence retrieval subtask. The results of our study point researchers to certain directions such as \revisedgiannis{
\begin{enumerate*}[label=(\roman*)]
\item \nikos{that} sentence sampling can be beneficial (\nikos{for example, in avoiding dataset imbalance and reducing computational complexity}), or
\item \nikos{that} working on the sentence retrieval step can lead to same performance improvement as in the case that one would work on the claim verification step \giannisrtwo{(when one considers the standard pipeline setting, where the task is divided into a series of three subtasks, see~\secref{sec:methods} for more details)}
\end{enumerate*}}.

\section{Background} 
In this section, we 
define the FEVER task and the problem it solves (\secref{sec:problem_definition}),
describe the way that the dataset is constructed (\secref{sec:dataset_construction}), and 
\revisedgiannis{present the subtasks of the FEVER task} (\secref{sec:preliminaries}).

\subsection{Problem Definition}
\label{sec:problem_definition}
The FEVER shared task provides a set of claims where each claim is a sentence \giannisrthree{the veracity of which} should be identified. The veracity of a claim should be based on (sentence-level) evidence provided by Wikipedia. For that, a set of pre-processed (from the year 2017) Wikipedia documents has been shared with the participants of the competition. A claim can be either {\tt{SUPPORTED}} or {\tt{REFUTED}}, assuming that correct evidence has been identified. In the case that there is not enough information in Wikipedia, the veracity of the claim should be assessed as {\tt{NOTENOUGHINFO} (NEI)}. The goal of the task is to return \nikos{for each claim either the {\tt{SUPPORTED}}/{\tt{REFUTED}} label along with the corresponding evidence or the {\tt{NEI}} label without evidence}.
In 16.82\% of the claims, more than one evidence sentences are needed to conclude about the veracity of the claim \revisedgiannis{(the evidence is from different pages/documents in 12.15\% of the claims)}. Two examples \giannisrthree{from} the FEVER dataset are illustrated in~\figref{fig:fever_examples}.

\begin{figure}[t]
\resizebox{0.8\columnwidth}{!}{%
\begin{minipage}{\textwidth}
    \parskip=0pt
    \begin{framed}
    \parskip=0pt
      \begin{description}

          \item[Claim:] Claire Danes is wedded to an actor from England.
         
          \item[\texttt{[wiki/Claire\_Danes]}] \textcolor{RoyalBlue}{She is married to actor Hugh Dancy}, with whom she has one child.

\item[\texttt{[wiki/Hugh\_Dancy]}] \textcolor{RoyalBlue}{Hugh Michael Horace Dancy} (born 19 June 1975) \textcolor{RoyalBlue}{is an English actor} and model. 
      \item[Verdict:] {\textcolor{OliveGreen}{\tt{SUPPORTED}}}

      \end{description}
    \parskip=0pt
\end{framed}
\parskip=0pt
    \begin{framed}

      \begin{description}
    \item[Claim:] Rogue appears in Canadian comic books.
         
          \item[\texttt{[wiki/Rogue\_(comics)]}] \textcolor{RoyalBlue}{Rogue is} a fictional superhero \textcolor{RoyalBlue}{appearing in American comic books} published by Marvel Comics, commonly in association with the X Men .

      \item[Verdict:] {\textcolor{BrickRed}{\tt{REFUTED}}}

      \end{description}
      
    \parskip=0pt
\end{framed}
\end{minipage}}
  \caption{Two examples from the FEVER dataset where evidence sentences should be selected from Wikipedia. In the example illustrated on top, the claim is {\textcolor{OliveGreen}{\tt{SUPPORTED}}} and the relevant information from Wikipedia is indicated in \textcolor{RoyalBlue}{blue} color. In the example illustrated in the bottom, the claim is {\textcolor{BrickRed}{\tt{REFUTED}}} by the evidence sentence.}
  \label{fig:fever_examples}  
\end{figure}

\subsection{Dataset Construction} \label{sec:dataset_construction}
The FEVER dataset includes 185,445 claims and the \nikos{numbers} of the examples per label (\ie {\tt{SUPPORTED}},  {\tt{REFUTED}}, {\tt{NEI}}) for training, development and test are presented in~\tabref{tab:dataset_statistics}. The FEVER dataset has been constructed in two phases:  \begin{enumerate*}[label=(\roman*)]
\item claim generation and\label{phase1}
\item claim labeling.\label{phase2}
\end{enumerate*} 

In total, 50 annotators have contributed in the process. In phase~\ref{phase1}, the annotators created claims from randomly chosen Wikipedia sentences. The claims should be sentences that include a single piece of information. The goal of the claim generation phase is to create claims that are not trivially verifiable (\ie too similar to the source) nor too complex. For that, hyperlinks have been included in the sentences in order for the annotators to incorporate external knowledge in a controlled way. Except for the original claims, the annotators created variations of the claims by, for example, paraphrasing, adding negation. For the claims that were the negated versions of the original claims, the authors have observed that only trivial negations were generated (\ie by adding only the word ``not''). To alleviate this issue, the annotation interface has been re-designed to highlight the ``not'' trivial negations.
In phase~\ref{phase2} of the dataset construction process, the annotators were asked to label the claims as {\tt{SUPPORTED}},  {\tt{REFUTED}} or {\tt{NEI}}. For the {\tt{SUPPORTED}} and  {\tt{REFUTED}} labels, the annotators also provided the sentences that have \nikos{been} used as evidences for supporting or refuting the veracity of the claim. For the {\tt{NEI}} label, only the label itself was provided since the annotator could not conclude whether the claim was supported or refuted based on the available Wikipedia sentences. Finally, to improve the quality of the provided dataset, 
\begin{enumerate*}[label=(\roman*)]
\item super-annotators checked randomly 1\% of the data,
\item the Fleiss $\kappa$ score~\cite{fleiss:71} \revisedgiannis{(which measures the inter-annotator agreement among a fixed number of annotators when assigning categorical labels to a number of instances)} for 4\%  of randomly selected claims has been calculated among five annotators, and
\item the authors have manually re-validated the quality of the constructed dataset (for 227 examples). 
\end{enumerate*} 

\subsection{FEVER Subtasks} \label{sec:preliminaries}
In the literature, the FEVER task has been mostly treated as a series of three subtasks, namely document retrieval, sentence retrieval and claim verification~\cite{thorne:18,nie:19,yoneda:18}. 
\begin{table}[ht]
\resizebox{0.4\columnwidth}{!}{%
\begin{tabular}{@{\extracolsep{4pt}}ccccccc@{}} 
 \toprule
  
\multicolumn{1}{c}{Split}&  \multicolumn{1}{c}{{\tt{SUPPORTED}}}& \multicolumn{1}{c}{{\tt{REFUTED}}}& \multicolumn{1}{c}{{\tt{NEI}}}  \\
 
 \midrule

 Train&80,035 & 29,775 & 35,639 \\
 Dev&6,666 & 6,666 & 6,666   \\
Test&6,666 & 6,666 & 6,666 \\
\bottomrule
\end{tabular}
}
 \caption{The statistics of the FEVER dataset as presented in~\citet{thorne:18b}.}
\label{tab:dataset_statistics}
 \end{table}
\giannisrthree{\subsubsection{Document Retrieval} \label{sec:preliminaries_doc_retrieval}
Document retrieval is the task that aims at matching a query (\ie claim in the context of FEVER) against a collection of unstructured documents (\ie Wikipedia in the context of FEVER) and returns the most relevant articles~\cite{chen:17} (see \secref{pipeline_models} for more details).} 
\giannisrthree{\subsubsection{Sentence Retrieval}
Given a query (\ie claim in the context of FEVER), the goal of sentence retrieval is to find the relevant sentences (\ie evidence sentences in the context of FEVER) out of a given document, or a set of documents, retrieved from the document retrieval step. 
\subsubsection{Claim Verification}
The claim verification task aims at verifying the veracity of a given claim (\ie {\tt{SUPPORTED}},  {\tt{REFUTED}} or {\tt{NEI}} as defined in \secref{sec:problem_definition}). In the context of FEVER, the veracity of  a claim is assessed by taking the retrieved evidence sentences from the sentence selection subtask into account. 
For a detailed description of the FEVER  subtasks see \secref{pipeline_models}.}
\begin{figure*}
\resizebox{.7\linewidth}{!}{
\centering				
\includegraphics[width=\textwidth]{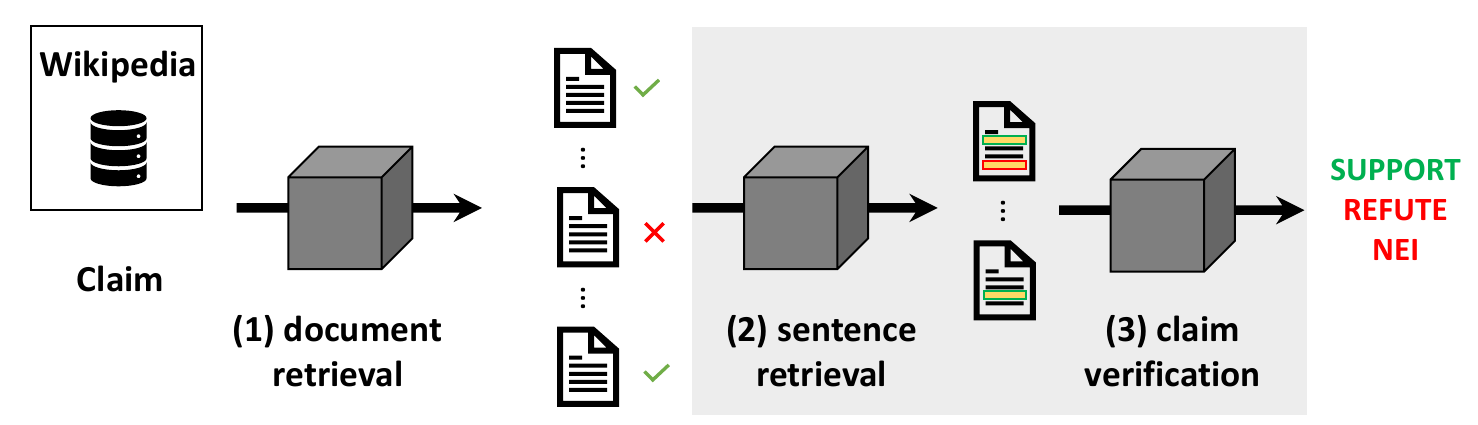}}
\caption{A three-step pipeline model for the FEVER task. It consists of three components, namely document retrieval, sentence retrieval and claim verification. The input to the first component is Wikipedia and a given claim sentence. The output of the first component is a set of Wikipedia documents related to the claim. The retrieved documents are fed as input to the sentence retrieval component and the output of that module is a set of sentences related to the claim from the input documents. Finally, the input to the claim verification component is the retrieved sentences from step (2) and the output is a label which indicates the veracity of the claim. Note that the claim is provided as input to every component of the pipeline system. The shaded box illustrates the fact that in several systems, steps (2) and (3) are performed in a joint setting.}
\label{fig:fever_pipeline}
\end{figure*}
\subsection{Three-Step Baseline Model}\label{sec:baseline_model}
Along with the FEVER dataset, \citet{thorne:18} provided a three-step pipeline model to solve the FEVER task \giannisrthree{(see~\secref{sec:preliminaries} for more details on the three subtasks)}. 
A graphical illustration of the three-step pipeline model is provided in~\figref{fig:fever_pipeline}. Most of the existing \giannisrtwo{studies} so far (see~\citet{zhao:20,zhong:20}) are also following this three-step pipeline approach; however, more complex architectures  have been proposed to solve the FEVER task in an end-to-end fashion~\cite{yin:18}.

\subsection{Evaluation}
In this subsection, we describe the evaluation metrics that are used for evaluating the performance in the different FEVER subtasks. Note that since the three subtasks are stacked the one on top of the other, higher performance in the downstream components (\eg document retrieval) leads to better performance on the upstream components (\eg sentence retrieval and claim verification). The organizers of the FEVER shared task~\cite{thorne:18b} have released a Github repository with the code of their evaluation module\footnote{\url{https://github.com/sheffieldnlp/fever-scorer}}.
\subsubsection{Document Retrieval}\label{sec:evaluation:document_retrieval}
The evaluation of the results for the subtask of document retrieval \nikos{is} based on the work of~\citet{thorne:18}, \giannisrthree{which uses two metrics}, namely, \textit{oracle accuracy} and \textit{fully supported}. The \textit{fully supported} metric indicates the number of claims for which the correct documents (\ie along with the corresponding evidences) have been fully retrieved by the document retrieval component. This metric \giannisrthree{only takes} into account the claims that are supported/refuted by evidences (\ie does not consider the {\tt{NEI}} class). The \textit{oracle accuracy} is the upper bound of \nikos{the} accuracy over all the three classes (\ie considers the claims of the {\tt{NEI}} class as correct).
\subsubsection{Sentence Retrieval}\label{sec:evaluation:sentence_retrieval}
The evaluation of this subtask is performed by using \textit{precision}, \textit{recall} and \textit{\Fone}~scores.
Specifically, the organizers of the shared task suggested the precision to count the number of the correct evidences retrieved by the sentence retrieval component with respect to the number of the predicted evidences for the supported/refuted claims. 
The recall has also been exploited for the supported/refuted claims. A claim is considered correct in the case that at least a complete evidence group is identified. Finally, the \Fone~score is calculated based on the aforementioned metrics.
\subsubsection{Claim Verification}\label{sec:evaluation:claimm_verification}
The evaluation of the claim verification subtask is based on the \textit{label accuracy} and the \textit{FEVER score} metrics. The label accuracy measures the accuracy of the label predictions (\ie {\tt{SUPPORTED}}, {\tt{REFUTED}} and {\tt{NEI}}) without taking the retrieved evidences into account. On the other hand, the FEVER score counts a claim as correct if a complete evidence group has been correctly identified (for the supported/refuted claims) as well as the corresponding label. Thus, the FEVER score is considered as a \giannisrthree{stricter evaluation metric than label accuracy} and it was the primary metric for ranking the systems on the leaderboard of the shared task.

\begin{sidewaystable*}

\resizebox{\columnwidth}{!}{%
\centering
\begin{tabular}{@{\extracolsep{4pt}}ccccccc||cccc||cccccccc@{}} 
 \toprule
 & \multicolumn{1}{c}{}   &  \multicolumn{5}{c}{Document Retrieval}&  \multicolumn{4}{c}{Sentence Retrieval} &   \multicolumn{6}{c}{Claim Verification} &  \multicolumn{2}{c}{Joint}\\
\cline{3-7}
\cline{8-11}
\cline{12-17}
\cline{18-19}
 & \multicolumn{1}{c}{Model} &  \multicolumn{3}{c}{Term-based} & \multicolumn{1}{c}{DRQA} & \multicolumn{1}{c}{Features}  & \multicolumn{1}{c}{TF-IDF} & \multicolumn{1}{c}{ESIM} & \multicolumn{1}{c}{LM}& \multicolumn{1}{c}{Other}  & \multicolumn{3}{c}{Neural} & \multicolumn{3}{c}{LM} & \multicolumn{1}{c}{Supervised}& \multicolumn{1}{c}{Generated} \\
 \cline{3-5}
 \cline{12-14}
  \cline{15-17}
  & \multicolumn{1}{c}{} &  \multicolumn{1}{c}{Mention} & \multicolumn{1}{c}{Keyword}& \multicolumn{1}{c}{Exact  Match} & \multicolumn{1}{c}{} & \multicolumn{1}{c}{} & \multicolumn{1}{c}{} & \multicolumn{1}{c}{} & \multicolumn{1}{c}{} & \multicolumn{1}{c}{}  & \multicolumn{1}{c}{ESIM}  & \multicolumn{1}{c}{LSTM/CNN} & \multicolumn{1}{c}{DA}& \multicolumn{1}{c}{Simple}  & \multicolumn{1}{c}{Graph} & \multicolumn{1}{c}{Seq2seq}\\
 \midrule
\parbox[c]{5mm}{\multirow{10}{*}{\rotatebox[origin=c]{90}{\parbox{1.0cm}{\centering 2018}}}}

 &\citet{hanselowski:18}&\cmark&&&&&&\cmark&&&\cmark&&&&&&& \\
&\citet{thorne:18} &&&&\cmark&&\cmark&&&&&&\cmark&&&&&\\
&\citet{yoneda:18} &\cmark&&&&\cmark&&&&\cmark&\cmark&&&&&&&\\
 &\citet{yin:18a} &&&&\cmark&&\cmark&&&&&\cmark&&&&&& \\
&\citet{hidey:18}  &\cmark&&&\cmark&\cmark&&\cmark&&&\cmark&&&&&&\cmark&\\
&\citet{chakrabarty:18}  &\cmark&&&&&\cmark&&&&&\cmark&&&&&&\\
 &\citet{malon:18} &\cmark&&&&&&&&&&\cmark&&&&&& \\
&\citet{luken:18} &&\cmark&&&&&&&\cmark&&&\cmark&&&&&\\
&\citet{yin:18}  &\cmark&&&&&&&&\cmark&&\cmark&&&&&\cmark&\\
&\citet{taniguchi:18}  &&&\cmark&&&\cmark&&&&&\cmark&&&&&&\\
 \midrule
\parbox[c]{5mm}{\multirow{8}{*}{\rotatebox[origin=c]{90}{\parbox{1.0cm}{\centering 2019}}}}

 &\citet{nie:19}  &&\cmark&&&&&\cmark&&&\cmark&&&&&&& \\
&\citet{nie:19a}  &&\cmark&&&&&&\cmark&&&&&\cmark&&&&\\
 &\citet{ma:19}  &&\cmark&&&&&&&\cmark&&\cmark&&&&&& \\
&\citet{zhou:19}  &\cmark&&&&&&\cmark&&&&&&&\cmark&&&\\
 &\citet{chernyavskiy:19}  &\cmark&&&&&\cmark&&&&&&&\cmark&&&& \\
&\citet{stammbach:19}   &\cmark&&&&&&\cmark&\cmark&&&&&\cmark&&&&\\
&\citet{jobanputra:19}   &&&&&&&&&&&&&\cmark&&&&\cmark\\

&\citet{tokala-etal-2019-attentivechecker}   &&&\cmark&&&&&&\cmark&&\cmark&&&&&&\\
 \midrule
\parbox[c]{5mm}{\multirow{13}{*}{\rotatebox[origin=c]{90}{\parbox{1.0cm}{\centering 2020}}}}

&  \citet{zhao:20} &\cmark&&&&&&&\cmark&&&&&&\cmark&&&\\
&\citet{liu:20} &\cmark&&&&&&&\cmark&&&&&&\cmark&&& \\
&\citet{soleimani:20} &\cmark&&&&&&&\cmark&&&&&\cmark&&&& \\
&\citet{zhong:20} &&\cmark&&&&&&\cmark&&&&&&\cmark&&&  \\
&\citet{portelli:20} &&\cmark&&&&\cmark&&&&&&&\cmark&&&& \\
&\citet{lee:20}   &&&&&&&&&&&&&\cmark&&&&\cmark\\
&\citet{lewis:20}   &&&&&&&&&&&&&&&\cmark&&\cmark\\
&\citet{lewis2020bart,petroni2020kilt}   &&&&&&&&&&&&&&&\cmark&&\cmark\\
&\citet{lewis:20,petroni2020kilt}   &&&&&&&&&&&&&&&\cmark&&\cmark\\
&\citet{nie:20}   &&\cmark&&&&&\cmark&&&\cmark&&&&&&\cmark&\\

&\citet{chen2020loren}   &\cmark&&&&&&&\cmark&&&&&\cmark&&&&\\

&\citet{ye2020coreferential}   &\cmark&&&&&&&\cmark&&&&&\cmark&&&&\\

&\citet{subramanian2020hierarchical}   &\cmark&&&&&&&\cmark&&&&&\cmark&&&&\\

\bottomrule
\end{tabular}
}
 \caption{Timeline with the \giannisrtwo{studies} that have been developed so far for the FEVER task, grouped based (i)  on the year and (ii) in a similar way to the one presented in~\secref{sec:methods}. LM stands for language model based approaches and the \cmark~symbol indicates whether a model uses a particular method. Note that most of the \giannisrtwo{studies} developed in 2019-2020 rely on pre-existing document retrieval components and the main focus is on the sentence retrieval and the claim verification components.}
\label{tab:timeline}
 \end{sidewaystable*}

\section{Methods}\label{sec:methods}
In this section, we describe the various methods that have been developed so far for solving the FEVER task. Most of the existing studies in the literature~\cite{hanselowski:18,zhong:20} divide the task into a series of three subtasks (\ie document retrieval, sentence selection and claim verification, see~\secref{pipeline_models} for a detailed description) similar to the baseline model as described in~\secref{sec:baseline_model}. However, there are some studies that merge the two subtasks of sentence selection and claim verification into one (see~\figref{fig:fever_pipeline}) mostly by exploiting multi-task learning architectures~\cite{yin:18,nie:20}. For a detailed description of these joint architectures, \nikos{we refer to}~\secref{sec:joint_models}.~\Tabref{tab:timeline} \nikos{presents} a timeline that summarizes the architectures developed so far for the FEVER task.
\subsection{Pipeline Models} \label{pipeline_models}
\subsubsection{Document Retrieval} \label{sec:pipeline_models:document_retrieval}
\giannisrthree{In this subsection, we describe the main methods that have been proposed for the document retrieval task.}
\nikos{The} input to the document retrieval step is Wikipedia and a given claim sentence. \giannisrthree{The output of this module is a set of Wikipedia documents relevant to the claim}. \\

\noindent\textbf{Preliminaries}\\
\revisedgiannis{\textit{DrQA:} Several approaches, which have been exploited to partly solve the FEVER task, rely on the DrQA component~\cite{chen:17} for retrieving relevant information from Wikipedia. The goal of DrQA is to answer questions on open-domain datasets such as Wikipedia. DrQA consists of two components 
\begin{enumerate*}[label=(\roman*)]
\item the \textit{document retriever}, which is responsible for identifying relevant articles, and \label{drqa:s1}
\item the \textit{document reader}, which is responsible for pointing to the start and end positions of the answers inside the document or a set of documents.
\end{enumerate*}
However, most of the existing literature on the FEVER task uses only component~\ref{drqa:s1} (\ie the document retriever) to collect relevant documents from Wikipedia (see~\secref{sec:problem_definition}). Specifically, the document retriever does not rely on machine learning methods. It calculates an inverted index lookup, computes the TF-IDF bag-of-words representations (bigrams) and scores the articles with the questions based on the aforementioned word vector representations.}\\

\noindent\textbf{Baseline}\\
\revisedgiannis{For this subtask, \citet{thorne:18} exploited the DrQA module, and used cosine similarity to obtain the $k$ most similar documents to the claim based on the TF-IDF word representation.}  \\

\noindent\textbf{Related work}\\
\textit{Mention-based methods:} \giannisrtwo{Several studies (see~\eg~\cite{hanselowski:18,chakrabarty:18}) have focused on the importance of named entities for the task of document retrieval. Thus, we have conducted a small scale analysis on the FEVER test set and we have observed that each claim contains more than $\sim$1.5 (on average) named entities and this number reaches up to 10 named entities for some claims. This indicates the importance of the named entities for the task of fact extraction and verification.}
\citet{hanselowski:18} proposed a mention-based approach to retrieve the relevant documents from Wikipedia for a given claim. This method consists of three components, namely,
\begin{enumerate*}[label=(\roman*)]
\item mention extraction,  \label{hanselowski:doc_retrieval_step1}
\item candidate article search, and \label{hanselowski:doc_retrieval_step2}
\item candidate filtering. \label{hanselowski:doc_retrieval_step3}
\end{enumerate*}
Component~\ref{hanselowski:doc_retrieval_step1} relies on a constituency parser, as developed in the work of~\citet{gardner:18}. Based on the parser, every noun phrase in a claim is considered a potential entity. \giannisrthree{In addition, all words before the main verb of the claim and the whole claim itself are also considered as potential entity mentions.} \nikos{Component~\ref{hanselowski:doc_retrieval_step2}, presented in the work of~\citet{hanselowski:18},} uses an external search API\footnote{\url{https://www.mediawiki.org/wiki/API:Main_page}} in order to match the potential entity mentions identified by component~\ref{hanselowski:doc_retrieval_step1} in the titles of Wikipedia articles. Component~\ref{hanselowski:doc_retrieval_step2} also returns some Wikipedia titles that are longer than the entity mentions. To deal with this case, component~\ref{hanselowski:doc_retrieval_step3} is responsible for stemming the Wikipedia title as well as the claim and discard all titles that are not part of the claim. The methodology of the work presented in~\citet{hanselowski:18} is also followed by~\citet{zhou:19,chernyavskiy:19,stammbach:19,zhao:20,liu:20,soleimani:20}, sometimes with minor modifications. Other works \nikos{that leverage the value of named entities are that} of~\citet{malon:18,chakrabarty:18,hidey:18,yin:18}. The work of~\citet{chakrabarty:18}, except for named-entity recognition uses the Google custom search API and dependency parsing as another task that improves the coverage of the retrieved documents. The \giannisrtwo{studies} of~\citet{chakrabarty:18} and~\citet{malon:18} also exploit disambiguation information, \eg whether the Wikipedia title refers to a ``film'' (\eg Titanic might refer to either the ship or the movie). \\ 

\noindent\textit{Keyword-based methods:}
The work of~\citet{nie:19} ranked at the first position at the FEVER competition. \nikos{They} presented a three-stage model that relies on Neural Semantic Matching Network (NSNM), \ie a variation of ESIM~\cite{chen:17b} (see~\secref{sec:pipeline_models:sentence_retrieval}). For the document retrieval, they exploit a keyword-matching approach that relies on exact matching (between the Wikipedia title and the spans of the claim), article elimination (\ie remove the first article \giannisrtwo{in the case that the claim starts with ``the'', ``an'' or ``a'' and apply the aforementioned matching scheme again -- note that this is different from stop word removal}) and singularization (\giannisrtwo{in the case  that there are no document titles returned then the claim is split into tokens and the aforementioned matching scheme is applied for every token}). Afterwards, all documents that do not contain disambiguative information (\eg ``band'', ``movie'') are added in the retrieved document list. The rest of the documents (\ie those with disambiguative information) are ranked and filtered out using NSNM and a threshold value. Several \giannisrtwo{studies}~\cite{ma:19,nie:19a,zhong:20,portelli:20} exploit the document retrieval module developed by~\citet{nie:19}. The work of~\citet{luken:18} is also designed to extract part-of-speech tags, dependencies, etc. by using the CoreNLP  parser~\cite{manning:14} for keyphrase identification. \\ 

\noindent\textit{Other methods:}
\giannisrtwo{The task of document retrieval is strongly connected to the task of information retrieval and standard schemes such as BM25 or cosine similarity in the embeddings space can be applied as baselines for retrieving the relevant documents.}
However, there are some methods that do not fall into any of the aforementioned categories. \giannisrthree{The work of~\citet{yin:18a} employ the module introduced in the baseline model~\cite{thorne:18} for the document retrieval step}. Similar to the baseline model, the work of~\citet{hidey:18} exploit DrQA along with hand-crafted features and neural methods for the document retrieval task.~\citet{yoneda:18} design hand-crafted features such as position, capitalization in the claim and train a logistic regression classifier. Unlike most of the \giannisrtwo{studies} on the document retrieval FEVER subtask that aim for high recall, the work of~\citet{taniguchi:18} aims for high precision using exact matching techniques. \revisedgiannis{Similar to this work, the work of~\citet{tokala-etal-2019-attentivechecker} relies on exact matching methods in order to reduce the number of available documents, but \nikos{it} also relies on a Bidirectional Attention Flow for Machine Comprehension
 (BIDAF) model~\cite{seo2016bidirectional} in order to rank the remaining documents.} In addition, as we observe in~\tabref{tab:timeline}, many of the \giannisrtwo{studies} that have been developed for the competition shared task (2018) focus on hand-crafted features (see categories ``Exact Match'', ``DrQA'', and ``Features''). However, this is not the case for more recent \giannisrtwo{studies} (2019-2020) that focus mostly on the sentence retrieval and claim verification components and use mention- and keyword-based approaches. \revisedgiannis{These \giannisrtwo{studies} are classified in separate categories in the timeline (\tabref{tab:timeline}), see columns ``Exact Match'', ``DrQA'', ``Features'', so that the reader can identify each component easily.}

\subsubsection{Sentence Retrieval}\label{sec:pipeline_models:sentence_retrieval}
In this subsection, we describe the main methods that have been proposed for the sentence retrieval component. The input to the sentence retrieval step is the Wikipedia documents retrieved from the previous component and the given claim sentence. Each Wikipedia document consists of sentences and \giannisrthree{the ones that are relevant to the claim}, the so-called evidences, are the output of the second component.\\

\noindent\textbf{Preliminaries}\\
\revisedgiannis{The tasks of sentence retrieval and claim verification are commonly framed as NLI problems (\ie are treated with methods originally developed to solve NLI tasks). This does not mean that the aforementioned methods are pre-trained on commonly used NLI datasets, but rather that the input of each subtask (\ie sentence retrieval and claim verification) is framed in such a way that traditional NLI methods can be used to resolve each of these two subtasks.}

\revisedgiannis{Assuming that we have two sentences, the \textit{hypothesis} and the \textit{premise} sentence, the goal of the NLI task is to determine whether the \textit{premise} sentence \textit{entails}, \textit{contradicts} or is \textit{neutral} to the \textit{hypothesis}. The most well-known datasets for NLI are the Stanford Natural Language Inference (SNLI) Corpus~\cite{bowman:15}, the Multi-Genre Natural Language Inference (MultiNLI) Corpus~\cite{williams:17}, and the cross-lingual NLI (XNLI) Corpus~\cite{conneau:18}. Several approaches have been proposed to solve the NLI tasks;  however, the neural models that have been mostly explored in the context of the FEVER task are the Decomposable Attention (DA)~\cite{parikh:16}, Enhanced 
Long Short-Term Memory (LSTM) for Natural Language Inference (ESIM)~\cite{chen:17b}, and Bidirectional Encoder Representations from Transformers (BERT)-based NLI~\cite{delvin:19}.\\ \\
\giannisrthree{\textit{ESIM:} \citet{chen:17b} rely on LSTM models to perform the NLI task. In particular, the model exploits the use of bidirectional LSTMs~\cite{hochreiter:97} (\ie on top of the word embeddings) to form representations of the premise and the hypothesis sentences. A soft-alignment layer that calculates attention weights 
is being used. In addition to the original representations, operations such as the difference and the element-wise product between the LSTM and the attended representations are calculated to model complex interactions. In the next layer, LSTMs are also exploited to construct the representation for the prediction layer. Finally, 
in the prediction layer, average and max pooling operations are used for the prediction of the final label.    \\ \\
\textit{DA:} The model has been proposed in the work of~\citet{parikh:16} and unlike the trend of using LSTMs, DA solely relies on word embeddings in order not to increase the complexity by $\mathcal{O}(d^2)$, where $d$ is the size of the hidden dimension. Specifically, DA consists of three components 
\begin{enumerate*}[label=(\roman*)]
\item the \textit{attention} step, which computes soft-alignment scores between the two sentences (\ie the premise and the hypothesis) similar to the method of~\citet{bahdanau:14},
\item the \textit{comparison} step, which applies a feed-forward neural network with a non-linearity between the aligned representations and 
\item the \textit{aggregation} step, which combines the information from previous steps via a summation operation to predict the final label.
\end{enumerate*}}
\\ \\
\textit{BERT:} Pre-trained Language Models (LM) have \giannisrtwo{been beneficial for a number of NLP tasks}. Examples include  ELMo (Embeddings from Language Models)~\cite{peters:18}, OpenAI GPT~\cite{radford:18} and BERT~\cite{delvin:19}. In the context of the FEVER task, several models~\cite{liu:20,soleimani:20} rely on the pre-trained BERT model. BERT relies on WordPiece \giannisrtwo{tokenization}~\cite{wu:16} and on Transformer networks~\cite{vaswani:17}. The input to the BERT model is either a single sentence or a pair of sentences encoded in a single sequence. The first token is \giannisrtwo{usually} the special token {\tt [CLS]}, which is used in classification tasks, and the sentences are separated by the special {\tt [SEP]} symbol. Two approaches have been proposed to pre-train the BERT model; specifically, 
\begin{enumerate*}[label=(\roman*)]
\item the Masked LM, where a percentage of random WordPiece tokens are masked and the goal is to predict the masked tokens, and 
\item the Next Sentence Prediction task, where the goal is to validate (0 or 1) whether the second sentence is the sequel of the first one.\label{nsp}
\end{enumerate*}
The pre-training task~\ref{nsp} has been shown to be extremely useful for downstream tasks such as Question Answering (QA) and NLI. For finetuning BERT for NLI, the sentence pair is separated by the {\tt [SEP]} symbol and the classification label \giannisrthree{(\eg entail, contradict or neutral)} is predicted on top of the {\tt [CLS]} symbol.}\\

\noindent\textbf{Baseline}\\
\revisedgiannis{For sentence selection, in the proposed three-step model,~\citet{thorne:18} obtained the most similar sentences from the retrieved documents (see previous subtask) by using either DrQA or unigram TF-IDF vectors. Moreover, they used a cut-off threshold tuned on the development set.}  \\

\noindent\textbf{Related work}\\
\textit{TF-IDF: }
For the sentence retrieval task, several pipeline methods in the literature rely on the sentence retrieval component of the baseline method~\cite{thorne:18}. Specifically, these methods~\cite{ chernyavskiy:19, portelli:20,taniguchi:18,yin:18a}, use a TF-IDF vector representation along with a cosine similarity function. However, there are some attempts that exploit additional representations such as \nikos{ELMo} embeddings~\cite{chakrabarty:18}. \\ \\
\textit{ESIM-Based:}
An important line of research~\cite{hanselowski:18,nie:19,zhou:19} for the sentence selection subtask includes the use of ESIM-based models~\cite{chen:17b}. Those \giannisrtwo{studies} formulate the sentence selection subtask as an NLI problem, where the claim is the ``premise'' sentence and the potential evidence sentence is a ``hypothesis'' sentence.~\citet{hanselowski:18} proposed a modified version of ESIM that during training receives as input the claim and the ground truth evidence sentences, as well as the claim with negative examples, randomly selected from the Wikipedia documents that the positive samples (\ie ground truth evidences) are coming from (\ie they sample randomly five sentences by not including the positive ones). The loss function used in this work is a hinge loss that receives as inputs the positive and the negative ranking scores (as pairs) from the ESIM model. At test time, the model computes the ranking score between the claim and each potential evidence sentence. It is also worth mentioning that the work of~\citet{zhou:19} exploits the evidences retrieved by the model of~\citet{hanselowski:18}. Similar to~\citet{hanselowski:18},~\citet{nie:19} use the same variation of ESIM called NSNM which has been exploited by the document retrieval component as well (see the keyword-based methods in~\secref{sec:pipeline_models:document_retrieval}).~\citet{nie:19} calculate the NSMN score between the claim and the evidence sentences. Afterwards threshold-based prediction is used to retain the highest scoring sentences. Unlike~\citet{hanselowski:18} that train a pairwise hinge loss,~\citet{nie:19} exploit a cross-entropy loss for training their model. \\ \\
\textit{Language Model Based:}
Similar to the ESIM-based methods, language model based methods~\cite{nie:19a,zhong:20,soleimani:20,liu:20,zhao:20,subramanian2020hierarchical} transform the sentence retrieval task to an NLI problem using pre-trained language models. The pre-trained language models are finetuned for the NLI task similar to the procedure described for the BERT-based model. It is however worth mentioning that the models developed for the sentence retrieval component do not rely only on BERT but also on RoBERTa~\cite{liu:19} and XLNet~\cite{yang:19}. For the language model based sentence retrieval two types of losses have been exploited:
\begin{enumerate*}[label=(\roman*)]
\item \nikos{the} pointwise loss, where a cross-entropy classifier is used to predict 0 or 1 (or a probability value), depending on whether the claim and the potential evidence sentence are related, and \label{pointwise_loss}
\item \nikos{the} pairwise loss, where the loss function takes as input a negative and a positive example. In that case, the positive example is the concatenation of the claim with an evidence sentence from the ground truth while a negative example is a concatenation between the claim and a negative example (\ie potential evidence, not included in the evidence set). That way, the model is learning to maximize the margin between the positive and the negative examples. \label{pairwise_loss}
\end{enumerate*}
For the loss of type~\ref{pointwise_loss}, several \giannisrtwo{studies} that use the BERT pre-trained model have been proposed~\cite{soleimani:20,nie:19a}. \giannisrthree{The work of~\citet{zhong:20}, which  also uses the pointwise loss, relies on RoBERTa and XLNet pre-trained models.} For \nikos{the} loss of type~\ref{pairwise_loss}, the proposed architectures rely only on the BERT pre-trained language model~\cite{soleimani:20,zhao:20,liu:20}. Due to the high number of negative examples with respect to the number of positive examples in the sentence retrieval subtask,~\citet{soleimani:20} proposed to use hard negative mining similar to ~\citet{schroff:15} to select more difficult examples (\ie those with the highest loss values). Note that training a pairwise loss is computationally more expensive than training a pointwise loss, since in the first case, one should consider all the combinations of positive-negative example pairs \giannisrtwo{(see \eg~\cite{liu:20})}. 
As we observe in~\tabref{tab:timeline}, most recent \giannisrtwo{studies} (\ie developed in 2019-2020) focus on developing language model based approaches.  
\\ \\
\textit{Other Methods:} 
\nikos{The two-step model of~\citet{stammbach:19} is able to combine both the ESIM-based and the language model based sentence retrieval components.} This work relies on the model of~\citet{nie:19} as a first component and uses a BERT-based model with two different sampling strategies to select negative examples.
\giannisrthree{Other alternative methods for sentence retrieval can be found in the following papers} ~\cite{luken:18,otto:18,yoneda:18,stammbach:19}.~\citet{luken:18} use  the root, the nouns and the named entities of the claim and construct a set of rules. For instance, if the named entities and the nouns are included in the sentence then the sentence is added in the evidence set. Similar to the work of~\citet{luken:18},~\citet{otto:18} also relies on nouns and named entities extracted from the claim using the spaCy NLP library~\cite{honnibal:15}. This work is able to directly retrieve evidences using the Solr indexer\footnote{\url{https://lucene.apache.org/solr/}} without relying on a document retrieval component.~\citet{yoneda:18} manually extract features such as the length of the sentences, whether the tokens of the sentence are included in the claim, etc. These features are fed into a logistic regression model. \revisedgiannis{Finally, the work of~\citet{tokala-etal-2019-attentivechecker} relies on the BIDAF model~\cite{seo2016bidirectional}, where the input is the claim and each candidate evidence sentence. The model output scores for each evidence sentence and, that way, is able to rank the candidate evidence sentences. \giannisrtwo{Note that passage retrieval techniques such as the one described in~\citet{karpukhin2020dense} or BM25 (or their combination) can also be used for sentence selection.}}

\subsubsection{Claim Verification}\label{sec:pipeline_models:claim_verification}
In general, most of the claim verification methods use neural model components. In the baseline method, the authors have exploited either a \giannisrthree{multi-layer perceptron (MLP)} neural model or a DA approach. In this section, the work on claim verification is divided into \begin{enumerate*}[label=(\roman*)]
\item ESIM-based architectures,
\item language model based approaches, and
\item other neural models.
\end{enumerate*} 
It is worth mentioning that most of the literature so far \giannisrthree{has} focused on improving the task of claim verification because the previous two subtasks (\ie document retrieval and sentence selection) have already attained quite good performance in terms of the recall evaluation metric, see~\secref{sec:results} and~\tabref{tab:timeline}.\\ 

\noindent\textbf{Preliminaries}\\
\revisedgiannis{For the claim verification subtask, researchers exploit \giannisrthree{techniques similar to the ones used} for the sentence retrieval subtask. For more details about the techniques that the claim verification models are based on can be found in~\secref{sec:pipeline_models:sentence_retrieval}.} \\

\noindent\textbf{Baseline}\\
\revisedgiannis{Two methods were developed for the claim verification component. First, an MLP was used by taking as input features the term frequencies of the claim and the evidence and the TF-IDF cosine similarity between them. Second, DA~\cite{parikh:16} -- which was described in~\secref{sec:pipeline_models:document_retrieval} -- has been used as a state-of-the-art system in NLI~\cite{bowman:15} (aka Recognizing Textual Entailment (RTE)).
It is worth mentioning that, for this step, evidences are needed in order to train the NLI component. However, this is not feasible for the {\tt{NEI}} labels, since there are no such evidence sentences in the training set. To circumvent this issue, two strategies have been explored in the baseline model: 
\begin{enumerate*}[label=(\roman*)]
\item sampling random sentences from Wikipedia, and 
\item sampling random sentences from the most similar documents as retrieved from the document retrieval component.
\end{enumerate*} }\\

\noindent\textbf{Related work}\\
\textit{ESIM-based:}
\citet{hanselowski:18} used an ESIM model for claim verification which has been modified to take as input multiple potential evidence sentences along with the given claim. They exploit the use of attention mechanisms, pooling operations and an MLP classifier to predict the relevant classes (\eg {\tt{SUPPORTED}}, {\tt{REFUTED}}, {\tt{NEI}}). The winning system of the FEVER task proposed by~\citet{nie:19} also relies on a modified version of ESIM called NSMN combined with additional features. This work exploits additional features such as WordNet embeddings (\ie antonyms, hyponyms), number embeddings and the scores from the previous subtasks.~\citet{yoneda:18} use the ESIM model where the claim with each potential evidence (and the associated Wikipedia article title) is considered independently. To aggregate the predictions for each evidence sentence with the claim,~\citet{yoneda:18} used an MLP classifier on top of the prediction score of each evidence sentence .  \\ \\
\textit{Language Model Based:}
Language models have also been  successfully applied to the claim verification subtask of FEVER. ~\citet{soleimani:20} formulated the problem of claim verification as an NLI task where the claim (premise) and the potential evidence sentence (hypothesis) are the inputs into a BERT-based language model. The evidences are independently considered against the claim and the final decision is made based on an aggregation rule similar to~\citet{malon:18} (\ie \nikos{the} default label is {\tt{NEI}} and, if there is a {\tt{SUPPORTED}} label, then the label of the claim is also {\tt{SUPPORTED}}). BERT-based models have also been adopted by multiple studies \eg \citet{nie:19a, stammbach:19, chernyavskiy:19, portelli:20,stammbach:20,subramanian2020hierarchical}.

\revisedgiannis{Graph Neural Network (GNN)-based Language Models:}~\citet{zhou:19} proposed a BERT-based method that makes use of GNNs~\cite{kipf:17}. By using GNNs, where evidences are nodes in a graph, they are able to exchange information between the nodes, \nikos{thereby} performing reasoning in order to obtain the final class label. Similar to the work of~\citet{zhou:19}, \nikos{\citet{liu:20} exploit} the use of kernel attention~\cite{xiong:17}, both at sentence and token level, to propagate information among the evidence nodes. A graph-based approach is also \nikos{explored} in the work of~\citet{zhong:20}, where, unlike previous \giannisrtwo{studies}, instead of using evidences as nodes in the graph, they construct the graph based on semantic roles (\eg verbs, arguments) as those extracted by an external library. Then, GNNs and graph attention mechanisms are used to combine and aggregate information among the graph nodes for the final prediction.~\citet{zhao:20} rely on a modified version of Transformers which is able to perform multi-hop reasoning even on long text sequences and combine information even along different documents. 

\revisedgiannis{Different from the previous \giannisrtwo{studies} that rely on GNNs, the paper of~\citet{ye2020coreferential} indicates that extracting coreference information from the text is important for claim verification. Specifically, in this paper, they rely on tasks such as entity masking in order to automatically exploit coreferential relations.} \revisedgiannis{Finally,~\citet{chen2020loren} extract the core parts of a claim, then they generate questions about these core parts and that way they are able to conclude about the veracity of the claim by performing reasoning on the question answering pairs.} \nikos{As we observe in~\tabref{tab:timeline},} most recent \giannisrtwo{studies} (\ie developed in 2019-2020) focus on developing language model based approaches. \revisedgiannis{\nikos{These \giannisrtwo{studies}} are classified in separate categories in the timeline (\tabref{tab:timeline}), see columns ``Simple'' (simple classification layer), ``Graph'', ``Seq2seq'', so that the reader can identify each component easily.}
\\ \\
\textit{Other Neural Models:}
\nikos{Alternative neural models have also been proposed~\cite{chakrabarty:18,yin:18a,luken:18,malon:18,taniguchi:18,otto:18}; these cannot be classified in any of the aforementioned categories.} Specifically,~\citet{chakrabarty:18} rely on bidirectional LSTMs and perform operations \nikos{combining the claim and evidence representations} (\eg element-wise product, concatenation) similar to~\citet{conneau:17}. Other studies use the DA model~\cite{luken:18,otto:18}, similar to the one that was exploited in the baseline model. 
Other methods such as Convolution Neural Networks (CNNs) with attention mechanisms~\cite{yin:18a,taniguchi:18} and transformer networks~\cite{malon:18} have also been used.
\revisedgiannis{Finally, the work of~\citet{tokala-etal-2019-attentivechecker} relies on the BIDAF model~\cite{seo2016bidirectional}, where the input is the claim and all the retrieved evidence sentences.} \revisedgiannis{These \giannisrtwo{studies} and the ESIM-based models are classified in separate categories in the timeline (\tabref{tab:timeline}), see columns ``ESIM'', ``LSTM/CNN'', ``DA''. \giannisrtwo{Note that QA (see \eg~\cite{kundu2018question,bajaj2016ms}) can also be used for claim verification similar to the work of~\citet{jobanputra:19}.}}

\subsection{Joint models}\label{sec:joint_models}
Unlike all the \giannisrthree{other} studies presented so far \nikos{in~\secref{sec:methods}, which consider} the FEVER subtasks in a pipeline setting, there has been a significant amount of work that handles the FEVER subtasks in a joint setting. The main motivation of joint methods is that in the pipeline setting, there are errors that are flowing from one component to the other, while in the case that two or more subtasks are considered together, decisions can be possibly corrected due to the interaction between the components. For instance,~\citet{yin:18} proposed the use of a multitask learning architecture with CNNs and attentive convolutions~\cite{yin:18a} in order to extract coarse-grained (\ie sentence-level) and fine-grained (\ie with attention over the words) sentence representations of \nikos{the} claim and evidences \nikos{so as} to perform the tasks of sentence selection and claim verification in a joint setting. Similar to this work,~\citet{hidey:18} train the sentence selection and claim verification subtasks in a multitask fashion. Specifically, they use the ESIM model for the representation of the claim and the evidence sentences, pointer networks~\cite{vinyals:15} for the sentence selection subtask and an MLP-based architecture for claim verification. A newer version of \nikos{this system, which uses adversarial instances to improve the performance,} has also been proposed in the work of~\citet{hidey:20}.~\citet{nie:20} perform an experimental study, where the NSNM model~\cite{nie:19} is compared in three different setups. In particular, a pipeline setting, a multitask setting and a newly introduced so-called ``compounded label set'' setting are being compared. This compounded label set setting is a combination of all the labels of the sentence selection and claim verification subtasks. 

\revisedgiannis{Unlike the previous line of research that trains the models in a multitask learning fashion, there are works that \nikos{address} some of the subtasks (\ie the retrieval steps) in an unsupervised manner. Specifically,~\citet{jobanputra:19,lee:20} formulate the problem as a \giannisrthree{masked
language modeling} task. In the work of~\citet{lee:20}, the last entity of the claim is masked out and the missing entity is filled in with a language model. That way, a new evidence is created (eliminating the need for a sentence retrieval module), which is then input together with the claim into an MLP to predict the claim verification label. Similar to that work,~\citet{jobanputra:19} also relies on language models and eliminates the sentence retrieval step by masking parts of the claim, \giannisrtwo{using a question answering module} and generating potential evidence sentences. Finally, the work of~\citet{lewis:20} is an end-to-end system that \giannisrthree{performs} the three steps at once. Specifically, they design a new retrieval system that is able to retrieve relevant documents and then generate text based on the retrieved documents. The model relies on a retriever that outputs a distribution of scores over the documents and a generator that takes also the previous generated tokens and the highest scoring documents into account. In the case of a classification task such as FEVER, where the goal is to predict a label based on a given claim, the sequence-to-sequence model (generator) has as goal (in the decoder part) to predict the tokens of the labels (\eg {\tt{SUPPORTED}}). This model has been exploited in other tasks such as question answering and question generation. \giannisrtwo{Similar to this model, the architecture proposed \giannisrthree{by}~\citet{lewis2020bart} also relies on sequence-to-sequence models that have been trained to reconstruct the original input after some sort of intentional document corruption. Similar to the model of~\citet{lewis:20}, it \giannisrthree{can} be used for sequence classification tasks. Note that although the model in~\cite{lewis2020bart} has not been initially used for the FEVER task, it can be used to address the FEVER task when it is paired with a state-of-the-art passage retriever~\cite{karpukhin2020dense} (which retrieves relevant passages for the classification task). It is worth stating that the models of \giannisrthree{M.~\citet{lewis2020bart} and P.~\citet{lewis:20}} are able to achieve state-of-the-art performance when knowledge from other Wikipedia-based intensive tasks (\eg entity linking, open domain question answering) is exploited (see~\cite{petroni2020kilt}).}}
\revisedgiannis{These \giannisrtwo{studies} are classified in separate categories in the timeline (\tabref{tab:timeline}), see columns ``Supervised'' and ``Generated''.}

\begin{table}
\resizebox{0.55\columnwidth}{!}{%
\begin{tabular}{@{\extracolsep{4pt}}ccccc@{}} 
 \toprule
& \multicolumn{1}{c}{Pre-calculated} & \multicolumn{1}{c}{}  &  \multicolumn{1}{c}{Fully}&  \multicolumn{1}{c}{Oracle} \\  
 \multicolumn{1}{c}{Model}& \multicolumn{1}{c}{Features}& \multicolumn{1}{c}{$\kappa$} &  \multicolumn{1}{c}{Supported} & \multicolumn{1}{c}{Accuracy}  \\
 \midrule
\citet{thorne:18}&\cmark&5   &55.30  &70.20     \\
\citet{yin:18}&\cmark&5   &89.63  &93.08     \\
\citet{hidey:18}&\cmark&5   &90.70  &-\\
\citet{hanselowski:18}&\cmark &7  &-  & 93.55     \\
\citet{chakrabarty:18}&\cmark &3  &\textbf{94.40}  & -     \\
\citet{zhou:19}&\cmark &7  &-  & 93.33      \\
 \citet{nie:19}&\cmark &5  &-  &92.42      \\
  
\citet{tokala-etal-2019-attentivechecker} & \cmark & 5  &-  &\textbf{94.05}      \\
\bottomrule
\end{tabular}
}
 \caption{Results of the document retrieval task in terms of the oracle accuracy and the fully supported evaluation metrics \giannisrtwo{(see~\secref{sec:evaluation:document_retrieval} for more details about the metrics)} in the dev set. The pre-calculated features column indicates whether a model uses external NLP tools or hand-crafted features. The symbol $\kappa$ is the number of the retrieved documents. The best performing models per column are highlighted in bold font. Missing results are not reported in the original papers.  }
\label{tab:results_doc_retrieval}
 \end{table}
\section{Results \& Discussion}\label{sec:results}
In this section, we describe the experimental results of the methods presented in \secref{sec:methods} and we compare their performance. Similar to the previous section, we present the results per subtask along with the corresponding discussion. \nikos{The} performance of the joint models is also presented in the corresponding subsections. 

\subsection{Document Retrieval}
In~\tabref{tab:results_doc_retrieval}, we present the results of the various document retrieval components that were extensively presented in~\secref{sec:pipeline_models:document_retrieval}. We evaluate the performance of the models based on the two commonly used evaluation metrics (\ie fully supported and oracle accuracy) for the document retrieval step of \nikos{the} FEVER task as introduced in the work of~\citet{thorne:18,thorne:18b} and presented in~\secref{sec:evaluation:sentence_retrieval}. In~\tabref{tab:results_doc_retrieval}, the pre-calculated features column indicates whether external NLP tools (\eg dependency parser) or hand-crafted features are exploited. In the $k$ column, the number of retrieved documents per claim is presented. \revisedgiannis{We report the $k$'s of each work that lead to best results or those that are used in the results of the pipeline systems.} The model of~\citet{thorne:18} is the baseline model as presented in~\secref{sec:baseline_model}. It is worth mentioning that in the various \giannisrtwo{studies}, there is \nikos{no} consistent report on the various metrics. The results are reported on the dev set since there is no ground truth data for this subtask on the test set. Specifically, the evaluation metrics on the competition platform\footnote{\url{https://competitions.codalab.org/competitions/18814}} for the test set assess the performance only of the subtasks two (\ie sentence retrieval) and three (\ie claim verification). Almost all of the systems presented in~\tabref{tab:results_doc_retrieval} rely either on mention- or keyword-based approaches, except for the baseline model that relies on TF-IDF features to obtain the most relevant documents. \nikos{The \giannisrtwo{studies} of~\citet{hanselowski:18} and~\citet{nie:19} are those that most of the recent neural methods rely upon (see~\secref{sec:pipeline_models:document_retrieval} for more details)}. The document retrieval presented in the work of~\citet{zhou:19} reproduces the results of the~\citet{hanselowski:18} model. In terms of oracle accuracy score, the model of~\citet{chakrabarty:18} is the best performing one; however, the models are not directly comparable since the oracle accuracy is measured based on a different number of retrieved documents. We observe that all the \nikos{models~\cite{yin:18,hidey:18,chakrabarty:18}, which rely on mention-based document retrieval,} achieve higher performance compared to the baseline model with respect to the fully supported evaluation metric. The same holds for the oracle accuracy evaluation metric. 
\revisedgiannis{The keyword-based model of~\citet{nie:19} also scores better compared to the baseline model.}
The gap between the keyword-based model of~\citet{nie:19} is relatively small in terms of the fully supported evaluation metric compared to the mention-based approaches. \revisedgiannis{The model of~\citet{tokala-etal-2019-attentivechecker} performs on par with the mention- and keyword-based approaches since it relies on exact match techniques \nikos{and the} BIDAF model~\cite{seo2016bidirectional}. We expect that there are no significant differences between the terms that are identified between the mention- and keyword-based (or the exact match) approaches since all of these approaches are able to identify the core parts of the document collections leading to similar performance as indicated in~\tabref{tab:results_doc_retrieval}. Thus, all of these systems form a good baseline for the document retrieval step. It is worth mentioning that in all of the presented models, the $k$ (\ie the number of retrieved documents) used in each study is relatively small (\ie 3-7) since otherwise the number of sentences for the next subtask (\ie sentence retrieval) would be relatively large. A large number of selected sentences has the immediate effect of significantly slowing down the training and inference at the sentence retrieval subtask.}

\begin{table*}
\resizebox{.8\textwidth}{!}{%

\begin{tabular}{@{\extracolsep{4pt}}ccccccccc@{}} 
 \toprule
 & \multicolumn{1}{c}{} & \multicolumn{1}{c}{Pre-calculated}   &  \multicolumn{3}{c}{Dev}&  \multicolumn{3}{c}{Test} \\ 
\cline{4-6}
\cline{7-9}
 & \multicolumn{1}{c}{Model}& \multicolumn{1}{c}{Features} &  \multicolumn{1}{c}{Precision} & \multicolumn{1}{c}{Recall} & \multicolumn{1}{c}{\Fone} & \multicolumn{1}{c}{Precision} & \multicolumn{1}{c}{Recall} & \multicolumn{1}{c}{\Fone}  \\

\midrule
\parbox[c]{5mm}{\multirow{3}{*}{\rotatebox[origin=c]{90}{\parbox{1.0cm}{\centering TF-\\IDF}}}}

 &\citet{thorne:18}&\cmark &-  & -      &17.20  & 11.28&47.87&18.26 \\
&\citet{taniguchi:18}&\cmark   &-  & -      &-  & 11.37&29.99&16.49\\
&\citet{chakrabarty:18}&\cmark   &-  & 78.04      &-  & 23.02& 75.89&35.33\\
 \midrule
\parbox[c]{5mm}{\multirow{3}{*}{\rotatebox[origin=c]{90}{\parbox{1.0cm}{\centering ESIM-\\based}}}}

 &\citet{hanselowski:18}&\xmark   &-  &87.10      &-     & 23.61&85.19&36.97 \\
 
 &\citet{zhou:19}&\xmark   &24.08  &86.72   &37.69 &  23.51 &84.66 & 36.80 \\
&\citet{nie:19}&\xmark   &36.49  & 86.79      &51.38  & 42.27&70.91&52.96\\

 \midrule
\parbox[c]{5mm}{\multirow{6}{*}{\rotatebox[origin=c]{90}{\parbox{1.0cm}{\centering Language\\models}}}}

&  \citet{nie:19a}&\xmark  & - &  -     &\textbf{76.87}     &-&-&\textbf{74.62}\\
&\citet{soleimani:20}$^{**}$ &\xmark  &38.18    &88.00  &53.25 &-&-&38.61 \\
&\citet{liu:20} &\xmark& 27.29& \textbf{94.37}       &42.34    &25.21&\textbf{87.47}&39.14  \\

&\citet{subramanian2020hierarchical} &\xmark& -& 90.50       &-     &-&-&-  \\
&\citet{zhong:20} &\xmark& 26.67& 87.64       &40.90     &25.63&85.57&39.45  \\
\midrule
\parbox[c]{5mm}{\multirow{4}{*}{\rotatebox[origin=c]{90}{\parbox{1.0cm}{\centering Other\\models}}}}
&\citet{luken:18} &\cmark  &\textbf{77.50}  & 52.30      &62.50    &\textbf{77.23}&47.12&58.53\\
&\citet{otto:18} &\cmark& -&  -      &-     &12.09&51.69&19.60  \\
&\citet{yoneda:18}$^{**}$ &\cmark& -& 84.54       &35.84     &22.16&82.84&34.97  \\
&\citet{stammbach:19}$^{**}$ &\xmark  &25.10    &89.80  &39.30 &-&-&- \\
 \midrule
\parbox[c]{5mm}{\multirow{3}{*}{\rotatebox[origin=c]{90}{\parbox{1.0
cm}{\centering Joint\\models}}}}
&\citet{yin:18}$^{*}$ &\xmark  &53.81  & 57.73      &50.59    &49.91&44.68&47.15\\
&\citet{hidey:18} &\xmark& -&  -      &-     &18.48&75.39&29.69  \\
&\citet{nie:20}&\xmark& -&  -      &-     &-&-&50.28  \\

\bottomrule
\end{tabular}
}
 \caption{Results of the sentence retrieval task in terms of the precision, recall, and \Fone~evaluation metrics in the dev and the test set. The pre-calculated features column indicates whether a model uses external NLP tools or hand-crafted features. The best performing models per column are highlighted in bold font. The single star symbol (*) denotes that the results of a model are reported on the dev and test sets defined in~\citet{thorne:18} (\ie 9,999 dev and 9,999 test instances) and not on the dev and test sets of the shared task~\cite{thorne:18b}. The double star symbol (**) indicates whether a model uses the title of the Wikipedia pages as external information. Missing results are not reported in the original papers. \revisedgiannis{It is worthwhile mentioning that all models that do not belong in the category ``Joint Models'' (see the vertical heading in the first column) are pipeline models.}}
\label{tab:results:sentence_retrieval}
 \end{table*}

\subsection{Sentence Retrieval}\label{sec:results_sentence_retrieval_section}
In~\tabref{tab:results:sentence_retrieval}, we present the results of the various sentence retrieval systems described in~\secref{sec:pipeline_models:sentence_retrieval}. 
The pre-calculated features column indicates whether the models use external NLP tools (\eg named entity recognizers to detect mentions) or hand-crafted features. We present results both on the dev and the test sets using the precision, recall and \Fone~evaluation metrics described in~\secref{sec:evaluation:sentence_retrieval}. 

In \tabref{tab:results:sentence_retrieval}, we observe that the different models optimize over different evaluation metrics (\eg precision, recall). For instance, the system of~\citet{luken:18} optimizes over the precision evaluation metric, the system of~\citet{nie:19a} optimizes over the \Fone~score, while most \giannisrtwo{studies} optimize over recall. This is because the recall metric measures the number of the correctly retrieved evidences over the total number of the ground truth evidences. This is of great importance since the core evaluation of the task (\ie the FEVER score) requires at least one correctly retrieved evidence group along with the correct label for the claim in order to evaluate a claim as correct. Thus, retrieving more evidence groups maximizes the chance of retrieving a correct evidence group out of the retrieved evidence groups. However, the organizers have imposed the restriction of taking into account only the five highest scoring evidence groups. This restriction alleviates the issue of returning the full set of evidence groups that would lead to the \nikos{problems} of  
\begin{enumerate*}[label=(\roman*)]
\item a perfect recall and 
\item transforming the FEVER score to the label accuracy metric. 
\end{enumerate*} 
Therefore, a high recall at the sentence retrieval subtask helps at increasing the performance of the model in terms of the FEVER score on the subtask of claim verification (next subtask in the pipeline).

\begin{table}[ht]
\resizebox{0.5\columnwidth}{!}{%

\begin{tabular}{@{\extracolsep{4pt}}ccccccc@{}} 
 \toprule
   & \multicolumn{3}{c}{Dev}   &  \multicolumn{3}{c}{Test}\\
\cline{2-4}
\cline{5-7}
  \multicolumn{1}{c}{Model}& \multicolumn{1}{c}{P@5}& \multicolumn{1}{c}{R@5} &  \multicolumn{1}{c}{\Fone@5} & \multicolumn{1}{c}{P@5}& \multicolumn{1}{c}{R@5} &  \multicolumn{1}{c}{\Fone@5}  \\

 \midrule

 Pointwise&27.66 & 95.91 & 42.94 &23.77 & 85.07 &37.15 \\
 Pairwise&27.29 & 94.37 & 42.34 &25.21 & 87.47 & 39.14  \\



\bottomrule
\end{tabular}
}
 \caption{The performance of a BERT-based model trained on the sentence retrieval task using the pointwise and the pairwise loss functions on the dev and the test sets (see the work of~\citet{liu:20}) in terms of Precision (P), Recall (R), and \Fone~scores. The results are reported on the 5 highly ranked evidence sentences (\ie @5). Note that in the original paper the results on the pointwise loss are not reported due to page limitations and the pointwise results are obtained from their Github codebase.}
\label{tab:pointwise_pairwise}
 \end{table}

In~\tabref{tab:results:sentence_retrieval} some of the results, especially in the dev set, are missing while for the test set, the results are available through the competition leaderboard. We observe that the ranking of the models (\ie which model performs better compared to another model) in terms of their performance remains the same for the dev and the test set. However, for most of the models, the performance decreases in the test set. In terms of the recall, the ESIM-based and the language model based models perform better compared to the rest of the models. Exceptions are the models of~\citet{yoneda:18} (\ie the system which has been ranked as second in the shared task and relies on hand-crafted features, external tools and a logistic regression classifier) and the model of~\citet{stammbach:19} (which is a combination of an ESIM-based and a BERT-based system). It is worth mentioning  the experiment of~\citet{liu:20} which indicates that using language models instead of ESIM-based for sentence retrieval leads to an improvement of 3 percentage points on the test set in terms of the recall evaluation metric and to 1 \giannisrtwo{percent} improvement on the claim verification subtask in terms of the FEVER score (this is not presented in~\tabref{tab:results:sentence_retrieval}). The two types of loss functions (\ie pointwise and pairwise, see the language model based part in~\secref{sec:pipeline_models:sentence_retrieval}) that have been exploited for the sentence retrieval task, have been more extensively studied in the work of~\citet{soleimani:20} and in the work of~\citet{liu:20}. The experimental study of~\citet{soleimani:20} suggests that there is a little variation in terms of recall between the pointwise and the pairwise models, even in the case that hard negative mining~\cite{schroff:15} is used in order to select more difficult instances. On the other hand, in the work of~\citet{liu:20} (see~\tabref{tab:pointwise_pairwise}), we observe that there is variation between the 
two losses on the dev and on the test set. \revisedgiannis{Moreover, we observe that the pointwise loss performs better on the dev set while it performs worse on the test; this suggests that the pointwise loss overfits faster (while tuning the parameters) on the dev set. We hypothesize that this is because in the pairwise loss all the pairs of positive and negative examples are used while in the pointwise loss only a ratio of negative examples is used (\ie five negative examples for each positive). The ratio in the pointwise loss is used since otherwise we would have a highly imbalanced dataset, which can negatively affect performance.
This has also been verified in our experiments, presented in~\secref{sec:experimental_study_intro}, where the performance of the pointwise loss decreases by 1 percentage point in terms of the FEVER score both on the dev and the test set.} The lack of a consistent conclusion from the two experimental studies (\ie the one of~\citet{soleimani:20} and the one of~\citet{liu:20}) suggests that the impact of the loss function on the sentence retrieval task needs further investigation.

\begin{table*}[ht]
\resizebox{.8\textwidth}{!}{%

\begin{tabular}{@{\extracolsep{4pt}}ccccccc@{}} 
 \toprule
 & \multicolumn{1}{c}{} & \multicolumn{1}{c}{Pre-calculated}   &  \multicolumn{2}{c}{Dev}&  \multicolumn{2}{c}{Test} \\  
\cline{4-5}
\cline{6-7}
 & \multicolumn{1}{c}{Model}& \multicolumn{1}{c}{Features} &  \multicolumn{1}{c}{Label Accuracy} & \multicolumn{1}{c}{FEVER} & \multicolumn{1}{c}{Label Accuracy} & \multicolumn{1}{c}{FEVER}  \\

 \midrule
\parbox[c]{5mm}{\multirow{3}{*}{\rotatebox[origin=c]{90}{\parbox{1.0cm}{\centering ESIM-\\based}}}}

 &\citet{hanselowski:18}&\xmark   &68.49  &64.74      &65.46     & 61.58 \\
 &\citet{yoneda:18}$^{**}$&\xmark   &69.66  &65.41   &67.62 &  62.52 \\
&\citet{nie:19}&\cmark   &69.72  & 66.49      &68.21  & 64.21\\

 \midrule
\parbox[c]{5mm}{\multirow{12}{*}{\rotatebox[origin=c]{90}{\parbox{1.3cm}{\centering Language\\models}}}}
&  \citet{nie:19a}&\xmark  & 75.12 &  70.18     &72.56     &67.26\\
&\citet{chernyavskiy:19}$^{**}$ &\xmark& -&  -       &71.72     &67.68  \\

&\citet{zhou:19} &\xmark& 74.84&  70.69       &71.60     &67.10  \\
&\citet{portelli:20} &\xmark& \textbf{84.33}&  -       &-     &-  \\
&\citet{soleimani:20} &\xmark  &74.59    &72.42  &71.86 &69.66 \\
&\citet{liu:20}$^{**}$ &\xmark& 78.29& 76.11       &74.07    &70.38  \\
&\citet{zhong:20} &\cmark& 79.16&  -       &76.85  &   70.60  \\
&\citet{zhao:20}$^{**}$ &\xmark& 78.05&  74.98       &72.39     &69.07  \\

&\citet{chen2020loren} &\cmark& 81.12 & \textbf{78.94}       &\textbf{76.91}     &73.43  \\

&\citet{ye2020coreferential} &\cmark& - & -       &75.96     &72.30  \\

&\citet{subramanian2020hierarchical} &\xmark& 75.77 &73.44       &74.64     &71.48  \\
&\citet{stammbach:20}$^{**}$ &\xmark& 77.48& 74.98       &76.60&    74.27  \\
 \midrule
\parbox[c]{5mm}{\multirow{7}{*}{\rotatebox[origin=c]{90}{\parbox{1.0cm}{\centering Other\\models}}}}
&\citet{thorne:18} &\cmark  &51.37  &31.27      &48.84 &27.45\\ 
&\citet{chakrabarty:18} &\xmark  &58.77    &50.83  &57.45 &49.06 \\
&\citet{luken:18} &\cmark  &44.70  & 43.90      &50.12    &43.42\\
&\citet{malon:18}$^{**}$ &\xmark& -&  58.44       &61.08     &57.36  \\
&\citet{taniguchi:18} &\xmark& -&  -       &47.13     &38.81  \\
&\citet{otto:18} &\xmark& -&  -       &54.15     &40.77  \\

&\citet{tokala-etal-2019-attentivechecker} &\xmark& -&  -       &69.98     &66.72  \\

 \midrule
\parbox[c]{5mm}{\multirow{8}{*}{\rotatebox[origin=c]{90}{\parbox{1.0cm}{\centering Joint\\models}}}}
&\citet{yin:18}$^{*}$ &\xmark  &78.90    &56.16  &75.99 &54.33 \\
&  \citet{hidey:18} &\xmark  & - &  -     &59.72     &49.94\\
& \citet{nie:20} &\xmark  &-  & -      &66.21    &62.69\\
& \citet{lee:20} &\cmark  &-  & -     &57.00     &-\\
& \citet{lewis:20} &\xmark  &74.50  & -      &72.50     &-\\
& \citet{lewis2020bart,petroni2020kilt} &\xmark  &-  & -      &-     &\textbf{86.74}\\
& \citet{lewis:20,petroni2020kilt} &\xmark  &-  & -      &-    &86.31\\
& \citet{hidey:20} &\cmark  &76.74  & 73.17      &72.47     &68.80\\



\bottomrule
\end{tabular}
}
 \caption{Results of the claim verification task in terms of the label accuracy and the FEVER score evaluation metrics in the dev and the test set. The pre-calculated features column indicates whether a model uses external NLP tools or hand-crafted features. The best performing models per column are highlighted in bold font. The single star symbol (*) denotes that the results of a model are reported on the dev and test sets defined in~\citet{thorne:18} (\ie 9,999 dev and 9,999 test instances) and not on the dev and test sets of the shared task~\cite{thorne:18b}. The double star symbol (**) indicates whether a model uses the title of the Wikipedia pages as external information. Missing results are not reported in the original papers. \revisedgiannis{All models that do not belong in the category ``Joint Models'' (see the vertical heading in the first column) are pipeline models.}}
\label{tab:results:claim_verification}
 \end{table*}

\subsection{Claim Verification}
In~\tabref{tab:results:claim_verification}, the results of the claim verification subtask are presented in terms of the label accuracy and the FEVER score evaluation metrics. As in the above tables, the models are grouped based on the way that the groups have been formulated in~\secref{sec:pipeline_models:claim_verification}. 

As we observe, the models have a better performance in the dev set compared to their performance on the test set. This is because the test set is blind and the number of submissions to Codalab is limited. Thus, the competition participants can only check the performance of their model in the test set by submitting the prediction file on the competition platform. On the other hand, the dev set is publicly available, and therefore, it is likely that some of the systems overfit on the dev set. Based on the results of~\tabref{tab:results:claim_verification}, the systems that use language models have better performance both in terms of label accuracy and FEVER score in the dev and test sets compared to the rest of the models. This is because pre-trained language models have a superiority over the rest of the methods, since they have been trained on large corpora and thus they already \nikos{incorporate} prior knowledge. The ESIM-based models, which are the three highly-ranked models of the shared task, are the second best performing group of models  in terms of both metrics, although there is a gap of 4-8 percentage points in terms of the label accuracy evaluation metric and 3-6 percentage points in terms of FEVER score. In addition, the joint model of~\citet{yin:18} performs well on the label accuracy (similar to the language model based approaches), however, the FEVER score drops dramatically due to the low recall of their model in the sentence retrieval task. Recall that in the task of sentence retrieval task in the work of~\citet{yin:18}, they optimize over the \Fone~score, which favors both precision and recall unlike most of the \giannisrtwo{studies} that optimize only over recall (see~\tabref{tab:results:sentence_retrieval}). Note that the presented results of~\citet{yin:18} are reported on the splits defined in the work of~\citet{thorne:18} and not on the splits of the shared task~\cite{thorne:18b}. In general, joint models have shown an improved performance in a number of tasks (e.g., entity-relation extraction~\cite{miwa:16,bekoulis:18b}, POS tagging-dependency parsing-chunking-semantic relatedness-textual entailment~\cite{hashimoto:17}) since the error propagation between the various sequential tasks is alleviated. However, this is not the case for the proposed joint architectures for the FEVER problem except for the model of~\citet{hidey:20}. We hypothesize that this is due to the fact that in the FEVER problem, there are no annotated (\ie gold) sentences for the {\tt{NEI}} class and thus the different strategies of selecting examples (\eg by randomly selecting sentences of the returned documents for that class) are not that beneficial. On the other hand, in the pipeline setting, a sentence retrieval model is trained on the sentences of the {\tt{SUPPORTED}} and {\tt{REFUTED}} classes and this model can later on be used to retrieve sentences that are exploited as potential evidence sentences along with the corresponding claim to train the model on the {\tt{NEI}} class for the claim verification subtask. 

Based on the results of~\tabref{tab:results:claim_verification}, \giannisrtwo{the models described in the work \giannisrthree{of M.~\citet{lewis2020bart} and P.~\citet{lewis:20}} and applied in the KILT benchmark~\cite{petroni2020kilt} are able to achieve state-of-the-art results although the retrieval step is unsupervised. This is because these models have been mapped to a fixed snapshot of Wikipedia along with other Wikipedia related tasks (five distinct tasks and 11 datasets) such as entity linking, open domain question answering. By handling similar \giannisrthree{knowledge intensive tasks} together can improve the performance for the FEVER claim verification task. It is worth stating that the model of~\citet{lewis:20} is not able to perform that well when it is not combined with KILT related tasks (due to the knowledge sharing between the various tasks). However, the results of the \giannisrthree{M.~\citet{lewis2020bart} and P.~\citet{lewis:20}} models in the KILT setting indicate that we can actually substitute the standard pipeline setting of the three subtasks for solving the FEVER problem.}
Moreover, \giannisrtwo{from the perspective of the pipeline models,} the model proposed in the work of~\citet{stammbach:20} performs best in terms of the FEVER score in the test set. 
\revisedgiannis{This is because this model relies on multiple combined modules on the downstream components (\ie document and sentence retrieval), but its main benefit comes from the fact that it relies on the GPT-3 pre-trained language model~\cite{gpt3}}.
\revisedgiannis{The models presented in the work of~\citet{zhong:20} and~\citet{chen2020loren} lead to the best performance in terms of label accuracy both in the dev and test sets. This is due to the fact that these models rely on external tools. The first one uses the semantic role labeling tool of AllenNLP\footnote{\url{https://demo.allennlp.org/semantic-role-labeling}} for constructing the graph of the claim and the evidence sentences. That way the graph neural network used in this work is able to take into account the structure of the semantic roles (due to the use of the external tool) instead of extracting that information from the raw claim and evidence sentences during training. The work of~\citet{chen2020loren} uses again AllenNLP tools to extract the central phrases in a claim.} The contribution of the semantic roles and central phrases is also evident \nikos{from that} other \giannisrtwo{studies} that rely on graph neural networks (see~\citet{zhou:19,liu:20}) achieve lower performance in terms of label accuracy and FEVER score. \revisedgiannis{The gap between the model of ~\citet{liu:20} and one that uses semantic roles~\cite{zhong:20} on the test set is smaller in terms of FEVER \nikos{score; we} hypothesize that this is because the former model is able to extract more relevant sentences (\ie higher recall) at the sentence retrieval subtask. \giannisrtwo{Although we cannot measure the statistical significance (\ie we do not have access to the test set), in this case, a benefit of two \giannisrthree{points} in the sentence retrieval step (\ie the model of~\citet{liu:20} scores 87.47\% in terms of the recall of the system 
 and the model of~\citet{zhong:20} scores 85.57\% -- recall is the most important metric in the sentence retrieval step) 
 leads to a two \giannisrthree{points} improvement in terms of the FEVER score in the claim verification step. 
 This is something that we observe in these particular systems; however, the performance benefit in the claim verification step may generally vary for different sentence retrieval and claim verification setups.} Moreover, \giannisrtwo{it} is also evident \nikos{that} the model of~\citet{chen2020loren} performs similar to the model of~\citet{zhong:20} in terms of label accuracy on the test set, while it performs three \giannisrthree{points} better in terms of FEVER score. We believe that this is because the model of~\citet{chen2020loren} relies on the sentence retrieval of~\citet{liu:20}.} We should note that the use of external tools \revisedgiannis{(\eg semantic role labeling)} is beneficial in a number of tasks (\eg entity and relation extraction where a dependency parser has been exploited to improve the relation extraction task, see~\citet{miwa:16}). However as presented in the work of~\citet{bekoulis:18b,bekoulis:18a}, the performance of a model can be significantly reduced when the external tool (\eg a parser) has been trained on data coming from different domains (\eg news data)  or languages (\eg English) and is applied on data from another domain (\eg biological data) or language (\eg Dutch). \nikos{In addition, the models of~\citet{soleimani:20,liu:20,zhong:20,zhao:20}, which optimize for high recall in the sentence retrieval task, obtain almost similar performance on the FEVER score on the test set (\ie a variation of 1 percentage point)}.
Moreover, the models of~\citet{liu:20,zhong:20,zhao:20} are far more complex compared to the plain BERT-based model of~\citet{soleimani:20}; thus, it is clear that the main benefit (compared to prior studies,~\eg~\citet{hanselowski:18}) comes from the pre-trained models. 

\section{Related tasks} \label{sec:related_tasks}
In this section, we present tasks that are strongly related to automated fact checking. Note that this is a non extensive list of related \nikos{tasks; however, this list serves as indication of problems related} to automated fact checking that can be tackled by similar methodologies. \giannisrtwo{The list of research areas that can help into tackling automated fact checking are detailed in~\citet{cazalens2018computational}.}\\

\noindent\textbf{Adversarial Examples for FEVER:}
The second version of the FEVER shared task has also been organized~\cite{thorne:19b}. The goal was to generate adversarial examples able to improve the performance of the FEVER task. Specifically, there were three phases, namely 
\begin{enumerate*}[label=(\roman*)]
\item Build-it, where the goal was to develop a system for solving the FEVER task,
\item Break-it, where the goal was to generate adversarial instances to fool the Builder system, and
\item Fix-it, where the goal was to combine the Builder system with the generated adversarial instances of the Breaker system in order to improve the performance of the model.
\end{enumerate*} 
It is worth mentioning that \nikos{the} authors of the shared task have also introduced metrics for evaluating the quality of the generated adversarial instances.~\citet{niewinski:19} submitted the winning solution for the Breakers system and they used a language model based architecture along with a targeted vocabulary for generating adversarial examples.\\

\noindent\textbf{Fake News Detection:}
Fake news detection is strongly related to automated fact checking and some of the previous \giannisrtwo{studies} consider automated fact checking as a constituent of fake news, \nikos{see, for example,} this review paper on fake news detection~\cite{oshikawa:20}. Fake news detection refers to the task of assessing the validity of full articles, claims or social media posts. Thus, the type of the input (\eg full article) depends on the dataset \nikos{and the} downstream application. It is common in applications \nikos{that assess} the validity of fake news (\eg posts, articles) to use fake/non-fake as prediction labels~\cite{perez:18}. However, in other applications, there are more fine-grained labeling schemes, since an article can be partially fake (see~\eg the work of~\citet{mitra:15}). For solving the fake news detection problem several approaches have been proposed ranging from feature-based models~\cite{conroy:15} to neural network architectures~\cite{rashkin:17,nguyen:19}. \revisedgiannis{Note that there are several surveys that have been proposed for related tasks to fact verification and specifically for fake news detection~\cite{zhou:20review,SAQUETE2020112943,Sharma:19,BONDIELLI201938}.}\\

\noindent\textbf{Rumour Detection:}
According to the work of~\citet{zubiaga:18}, a rumour is a statement, which has not been officially verified at the time that the statement is posted. Specifically, in this task, binary classifiers are usually used to predict whether a statement is a rumour or not. For instance, one can see the PHEME dataset~\cite{zubiaga:16}, where the input are threads of tweets from Twitter users and the goal is to classify them as true or false (\ie rumour, non-rumour). The use of the interactions among the users' profiles has also been exploited in the literature (see~\citet{do:19}) to improve the model performance on the rumour detection task.\\

\noindent\textbf{Stance Detection:}
Stance detection is a closely related problem to fact checking and according to the study of~\citet{zubiaga:18}, stance detection could be used after the rumour detection module in rumour classification systems. Specifically, the rumour detector is responsible to identify whether a statement is rumour or not, while the stance detector is responsible for identifying the stance of the author of a text against a given statement (\eg {\tt{FAVOR}}, {\tt{AGAINST}}) that has been classified as rumour~\cite{kuccuk:20}.\\

\section{Related Datasets}\label{sec:datasets}
Similar to the FEVER dataset, several other datasets have been proposed for verifying the truthfulness of claims in several contexts. \giannisrtwo{The techniques that are being used in each of these datasets are not directly comparable since some of these tasks are related but the same models cannot be applied out-of-the-box to each other task. More information about the baseline methods are provided in the description of each dataset.}\\

\noindent\textbf{SCIFACT:} Recently, a dataset \nikos{that follows the paradigm of the FEVER dataset in the context of validating scientific claims} has been proposed by~\citet{wadden:20}. Specifically, in that work, ~\citet{wadden:20} constructed a set of 1,409 scientific claims and the veracity of each claim was evaluated against a corpus of 5,183 scientific abstracts. The baseline model proposed for solving this task includes a pipeline of three subtasks similar to FEVER, namely, abstract retrieval (similar to document retrieval), rationale selection (similar to sentence selection) and label prediction (similar to claim verification). \giannisrtwo{Similar to the work of ~\citet{wadden:20}, ~\citet{kotonya-toni-2020-explainable} construct a dataset from the public health domain (called PUBHEALTH), where specific expertise (\eg from epidemiology) is required for predicting the veracity of a given claim. In~\citet{Grabitz172940}, citation networks have been exploited for verifying the veracity of scientific claims.}\\

\noindent\textbf{LIAR:} The LIAR dataset has been introduced in the work of~\citet{wang:17} for fake news detection. In particular, the author of this work created a dataset \nikos{consisting} of 12.8k statements manually annotated from the POLITIFACT.COM website. Unlike the FEVER \nikos{dataset, which} has been constructed using only Wikipedia documents, the problem described in the LIAR dataset is closely related to fake news detection since the dataset is constructed using only news content (\eg tweets, interviews, Facebook posts). An instance consists of a statement, the person who made the statement, and the context (\eg president elections). The label of the instance falls into one of the six predefined fine-grained classes (false, barely-true, etc.). The baseline model proposed in that work consists of a combined architecture of CNNs and bidirectional LSTMs to predict the label given the statement and the metadata.\\

\noindent\textbf{LIAR-PLUS:} A variation of the original LIAR dataset is the dataset introduced in the work of~\citet{alhindi:18}. Unlike the LIAR dataset, the LIAR-PLUS dataset extracts automatically the justification for each statement. To do so, the authors of the LIAR-PLUS select the summary section of the articles or last sentences (if there is no summary section) as justifications. By exploiting this information the model performance increases for all the examined architectures.\\

\noindent\textbf{TABFACT:} All the datasets presented so far rely on extracting information from raw text. However, none of these studies consider structured or semi-structured information.~\citet{chen:19} introduced a dataset called TABFACT, which includes 117,854 manually labeled claims based on 16,753 tables from Wikipedia. The goal of the task is to predict the veracity of the claim and two labels have introduced for that (\ie {\tt{ENTAILED}}, {\tt{REFUTED}}). The challenge in this dataset is that it is not straightforward to extract information from a Wikipedia table. The authors of this work have proposed a Transformer- and a BERT-based solution to solve the problem. \\

\noindent\textbf{FTFY:} The FTFY dataset proposed in the work of~\citet{hidey:19} contains contrastive claims of Reddit posts. Specifically, the authors of this work crawled posts from Reddit that received ``Fixed That For You'' (FTFY) responses. These responses are edits of the original post where the person who responds modifies part of the original comment. They propose a methodology for automatically generating pairs of contrastive claims using a sequence-to-sequence model, which are not trivially contrastive (\ie not negation of the original claim). Recall that this was also an issue that the authors of the FEVER dataset tried to alleviate, as described in~\secref{sec:dataset_construction}. \\

\noindent\textbf{MultiFC:} Unlike the FEVER \nikos{dataset, which} relies on claims generated from Wikipedia, the MultiFC dataset \nikos{proposed by~\citet{augenstein:19}} relies on 26 fact checking websites. In particular, they constructed a dataset of 34,918 claims. The claims \nikos{were} crawled from various domains, where each domain has a different number of \nikos{labels; this is also the challenging aspect} of this dataset. In this work, they also propose a multi-task learning approach, which takes into account the relation between the labels of the different domains.  \\

\noindent\textbf{FakeCovid:} A recently constructed dataset inspired by the COVID-19 pandemic is the FakeCovid~\cite{shahi:20}. This dataset includes 5,182 news articles related to COVID-19, coming from 92 fact checking websites in 40 languages. Unlike the aforementioned datasets, this one exploits the use of multilingual sources. The authors also provided a BERT-based classification model for solving the task.\\

\noindent\textbf{Others:} Several other datasets have been proposed in the context of fact checking and the related to that tasks, such as fake news detection, rumour detection, etc. A non extensive list of papers includes PHEME~\cite{zubiaga:16}, SOME-LIKE-IT-HOAX~\cite{tacchini:17}, SYMMETRIC FEVER~\cite{schuster:19}, CLEF-2019 tasks 1 \& 2~\cite{atanasova:19,hasanain:19}, CLIMATE-FEVER~\cite{diggelmann2020climate} \giannisrtwo{, VITAMINC~\cite{schuster2021get} and Real World search-engine-based claims~\cite{thorne2021evidence}}. \nikos{Additional datasets have introduced in the \giannisrtwo{studies} of~\citet{popat:16,derczynski:17,perez:18,shu:20}}.
\revisedgiannis{Note also that the FEVER dataset has been included in the ERASER~\cite{deyoung:20} and KILT~\cite{petroni2020kilt} benchmarks. }

\section{Competitions}
In this section, we list some of the most recent \giannisrthree{shared tasks/challenges} related to the FEVER task.
In particular, along with the FEVER competition, several other competitions have been organized, which are related to fact checking applications (\eg fake news, fact checking based on table data (see TABFACT in~\secref{sec:datasets})). Specifically, in the context of FEVER, a second shared task (\textbf{FEVER 2.0})~\cite{thorne:19b} has been organized and the goal is to define Builders (systems that solve the first task), Breakers (generate adversarial instances to break prior methods that solve the FEVER problem) and Fixers (improving systems by exploiting adversarial instances), as also described in~\secref{sec:related_tasks}. Another competition has been organized for the newly developed dataset~\textbf{SCIFACT}~\cite{wadden:20}\footnote{\url{https://scifact.apps.allenai.org/leaderboard}}. In this competition, the goal is to identify the validity of a claim based on scientific abstracts. \textbf{CLEF 2020 CheckThat!} is a competition \giannisrthree{that has been running for} the last 3 years (\ie since 2018) \nikos{and, in its last edition,} the goal for Task 1 was to rank claims (of a political debate) based on whether these claims are interesting to be annotated and for Task 2 was to rank the evidence and check the veracity of the claims (in an Arabic dataset).  
\nikos{A competition has also been organized\footnote{\url{https://competitions.codalab.org/competitions/21611}}~\cite{chen:19} based on the \textbf{TABFACT} dataset (described in~\secref{sec:datasets}), where the goal is to verify the validity of a claim based on Wikipedia tables}. In the same line, the shared task SemEval 2021 (Task 9) \textbf{SEM-TAB-FACT}\footnote{\url{https://sites.google.com/view/sem-tab-facts/}} aims at identifying whether a table supports a given claim and providing also the required evidence for that.  
The \textbf{Fake News Challenge}\footnote{\url{http://www.fakenewschallenge.org/}} has been organized in 2017 and attracted a lot of attention (\ie 50 participants). The goal was to identify the stance (\ie agree, discuss, disagree, be unrelated) of a specific article with respect to a headline. 

\section{Explainable Fact Checking Models}\label{explainability}
Although explainable models is an area of increasing interest in NLP~\cite{wallace:19,liu:19b}, there are only few studies that focus on explaining the outcome of fact checking models. Specifically,~\citet{atanasova:20} exploit the use of a joint architecture that simultaneously generates explanations using the extractive summarization model of \citet{liu:19c} and predicts the veracity of a claim using a classification layer. Similar to the previous work, the work of~\citet{stammbach:20} relies also on abstractive summarization techniques for generating explanations. \revisedgiannis{Specifically, this work, which relies on GPT-3~\cite{gpt3}, generates explanations based on the FEVER dataset. The model generates a summary based on the context (\ie evidences) and the claim. This summary has also been used as an input to the FEVER system (instead of the evidence sentences) and the system performed well on the dev set. This indicates that the summarized context (\ie evidences) is of high quality. \nikos{Along the same line,~\citet{glockner-etal-2020-think} proposed a new framework that is able to automatically extract rationales.} In particular, they also exploit the rationales to solve the task at hand (\ie they also demonstrate the effectiveness of their method in the FEVER dataset). The rationales that lead to best performance obtain a high ranking (\ie score). }  In another work,~\citet{ahmadi:19} propose the use of a set of rules on knowledge graphs (known for their structured representation of information, \ie entities and their corresponding relations) to extract interpretable results. Unlike the aforementioned studies that propose new methods for generating explanations, in the benchmark study of~\citet{deyoung:20}, they define different metrics for evaluating the quality of alignment between human and machine generated rationales.~\citet{nadeem:19} developed an end-to-end system for fact checking that is able to provide explanations based on the stance scores. Specifically, the sentences with the highest stance scores are highlighted in the user interface and provided as the explanations of the model. \revisedgiannis{The work of~\citet{paranjape-etal-2020-information} has identified that there should be an equilibrium between the level of explanation conciseness and the task accuracy. Experimental results indicate that retaining a trade-off between these two tasks leads to performance improvement on a standard benchmark (ERASER~\cite{deyoung:20}), as well as agreement with human generated explanations.}

\section{Experimental Study}
\label{sec:experimental_study_intro}
 
Inspired by~\tabref{tab:pointwise_pairwise} presented in~\secref{sec:results_sentence_retrieval_section}, we perform an analysis on the various loss functions for the sentence retrieval subtask. This is because based on the results presented in~\tabref{tab:pointwise_pairwise}, a concrete conclusion is missing about the type of loss function that is appropriate for sentence retrieval. As we observe in~\tabref{tab:pointwise_pairwise}, the pointwise loss function performs better on the dev set, while the pairwise loss function performs better on the test set  \revisedgiannis{and the rationale for this is explained at the end of \secref{sec:results_sentence_retrieval_section}}. Another motivation for this experimental study is that the performance of the loss functions used in the sentence retrieval task of the FEVER dataset is not a well-studied problem. Although, as analyzed in~\secref{sec:pipeline_models:sentence_retrieval}, several methods have been proposed for the sentence retrieval task, none of these studies \nikos{focuses} on experimenting with different ranking criteria. To investigate further the benefit of the examined loss functions, we conduct experiments both on the sentence retrieval and claim verification subtasks. 

\subsection{Methods}
We present the various loss functions that we have investigated for the sentence retrieval task. We have exploited loss functions commonly used in computer vision and document ranking. For representing the input sentences, we use a BERT-based model similar to previous work~\cite{liu:20}.\\

\noindent\textbf{Pointwise:} In this setting, we use the cross-entropy loss and the input to our model is the claim along with an evidence sentence. The goal of the sentence retrieval component paired with the pointwise loss is to predict whether a candidate evidence sentence is an evidence or not for a given claim. Thus, the problem of sentence retrieval is framed as a binary classification task.\\

\noindent\textbf{Pairwise:}
In our work, we also exploit the pairwise loss, where the goal is to maximize the margin between positive-negative examples. We use the pairwise loss similar to~\citet{wu:17}. The pairwise loss is: 
$
    \mathcal{L}^{pairwise}(p,n) = [- y_{ij}  (f(x_p)-f(x_n)) + m ]_+,
    \label{eq:pair}
$
\nikos{where} $y_{ij}\in\{-1,1\}$, $f(x)$ is the representation from the BERT-based model, $m$ is the margin and the indices $p$ and $n$ indicate a pair of a positive and a negative example.  
In order to obtain a claim aware representation of the instances, we concatenate the claim with the corresponding evidence. \\

\noindent\textbf{Triplet:}
Unlike the pairwise loss that considers only pairs of positive and negative examples, the triplet loss~\cite{wu:17} uses triplets of training instances. Given an anchor sample $a$, the goal is the distance $D_{ij} = \|f(x_i) -  f(x_j) \|_2$ to be greater between the anchor and a negative example than the distance between the anchor and a positive example. The triplet loss is:
$
    \mathcal{L}^{triplet}(a,p,n) = [ D_{ap}^2 - D_{an}^2 + m]_+,
    \label{eq:triplet}
$
\nikos{where} $m$ is the margin and the indices $a$, $p$ and $n$ indicate the triplet of the anchor, a positive and a negative example. As anchor we use the claim, \nikos{and} similar to the pairwise loss, we concatenate the claim with the corresponding evidence for positive-negative examples.\\

\noindent\textbf{Cosine:}
For the cosine loss, we exploit positive and negative samples by using the formula:
$\mathcal{L}^{cos}(p,n) = y_{ij} (1 - \cos(f(x_p), f(x_n))) +  (1-y_{ij}) [(\cos(f(x_p), f(x_n)) - m)]_+,
\label{eq:cosine}
$
\nikos{where} $y_{ij}\in\{0, 1\}$ and $\cos$ indicates the cosine distance between positive-negative samples.\\

\noindent\textbf{Angular:}
The angular loss~\cite{wang:17_angular} uses triplets of instances while imposing angular constraints between the examples of the triplet. The formula is given by:
$\mathcal{L}^{ang}(a,p,n) = [ D_{ap}^2 - 4\tan^{2}rD_{nc}^2 ]_+,
\label{eq:angular}
$\nikos{where} $f(x_c)= (f(x_a) - f(x_p)) / 2$ and $r$ is a fixed margin (angle). \\

\noindent\textbf{Evidence-Aware:}
Unlike all the aforementioned loss functions, we propose an evidence-aware model that relies on transformers, similar to~\citet{pobrotyn:20}. This model exploits the use of self-attention over the potential evidence sentences. Unlike
\begin{enumerate*}[label=(\roman*)]
\item the pointwise loss that does not take into account the relations between the evidence sentences and
\item the distance-based losses (\eg pairwise) that considers only pairs of sentences  
\end{enumerate*}, the evidence-aware model considers subsets of evidence sentences simultaneously at the training phase. Specifically, the input to the evidence-aware model is a list of BERT-based representations of the evidence sentences.  Thus, the model is able to reason and rank the evidence sentences while taking into account all the other evidence sentences in the list. We also exploit a binary cross-entropy loss similar to the one presented in the case of the pointwise loss, framing again the problem as a binary classification task.

\subsection{Setup}
We evaluate our models on the FEVER dataset~\cite{thorne:18} presented in~\tabref{tab:dataset_statistics}. We use the BERT-based model~\cite{delvin:19} in all of our experiments to guarantee a fair comparison among the various loss functions. The input to our sentence retrieval component is the output of the document retrieval task presented in the work of~\citet{hanselowski:18} and used also in the work of~\citet{liu:20}. For the conducted experiments in the sentence retrieval task, in all the loss functions except for the evidence-aware one, we present results using all the potential evidence sentences as retrieved from the document retrieval step. For the evidence-aware model, we conduct experiments using either 5 or 10 negative examples per positive instance during training. In addition, the overall (positive and negative) maximum number of instances that are kept is 20. This is because, unlike the other models \nikos{where} the evidences are considered individually or in pairs, in the evidence-aware model, we have to limit the instances that are considered simultaneously in the list. 
We experiment also with a limited number of instances in the other settings to have a fair comparison \nikos{among} the different setups. For the distance-based losses (\eg triplet, pairwise), we conduct additional experiments only in the best performing model when all instances are included (\ie the pairwise loss). We also present results on the claim verification task with all of the examined architectures. For the claim verification step, we use the model of~\citet{liu:20}. We evaluate the performance of our models using the official evaluation metrics for sentence retrieval (precision, recall and \Fone~using the 5 highly ranked evidence sentences) and claim verification (label accuracy and FEVER score) in the dev and test sets. 

\label{sec:experimental_study}
\begin{table*}
\resizebox{\textwidth}{!}{%

\begin{tabular}{@{\extracolsep{4pt}}ccccccccccccc@{}} 
 \toprule
 \multicolumn{1}{c}{} & \multicolumn{1}{c}{\# Negative} &  \multicolumn{1}{c}{\# Max}  &  \multicolumn{5}{c}{Dev}&  \multicolumn{5}{c}{Test} \\ 
\cline{4-8}
\cline{9-13}
\multicolumn{1}{c}{Loss}& \multicolumn{1}{c}{Examples} & \multicolumn{1}{c}{Instances} &  \multicolumn{1}{c}{P@5} & \multicolumn{1}{c}{R@5} & \multicolumn{1}{c}{\Fone@5}& \multicolumn{1}{c}{LA}& \multicolumn{1}{c}{FEVER} & \multicolumn{1}{c}{P@5} & \multicolumn{1}{c}{R@5} & \multicolumn{1}{c}{\Fone@5}& \multicolumn{1}{c}{LA}& \multicolumn{1}{c}{FEVER}  \\
\midrule

Angular&\cmark & \cmark &26.90  & 93.93 &41.82 &77.22 & 74.81   & 24.36&86.14&37.98 & 72.30  & 68.30\\
Cosine&\cmark &\cmark  &27.02  & 93.85       &41.96 & 77.50 & 75.10 & \textbf{24.83}&86.73&\textbf{38.61} & 72.49 & 68.81\\
Triplet&\cmark& \cmark  &26.99  & 94.24      &41.96 & 77.51 & 75.32  & 24.74& \textbf{86.86}&38.51 & 72.76 & 69.31\\
 \midrule
\parbox[c]{15mm}{\multirow{5}{*}{{\parbox{1.8cm}{\centering Pairwise}}}}
&\cmark&\cmark   &26.88  & 93.90      &41.79 & 78.05 & 75.61  & 24.44& 86.17&38.08 & 72.92&69.34\\
 &5& \cmark  &26.76  & 93.23      &41.58 &77.21&74.74  & 24.53& 85.90&38.17&72.05&68.22\\
 &10& \cmark  &26.77  & 92.99      &41.57 & 77.58 & 75.04 & 24.62& 86.15&38.29 & 72.65 &68.93\\
 &5& 20  &27.11  & 94.13      &42.10  & 77.53& 75.37&24.75 &86.67&38.51&72.87&69.25\\
 &10&20   &27.09  & 94.40      &42.10 & 78.05 & 75.79 &24.74& 86.84&38.51 &\textbf{73.02} & 69.38\\

 \midrule
 \parbox[c]{15mm}{\multirow{3}{*}{{\parbox{1.8cm}{\centering Pointwise}}}}
  &\cmark & \cmark &25.77  &91.96      &40.26     &77.94&75.12&22.28&82.61&35.01&71.63&67.63 \\
 &5 & \cmark   & 27.74      &95.93  &43.04& 78.43 & 76.71 &23.99&85.67&37.48 & 72.54 & 68.71\\
  &5 & 20   & 27.39      &95.25  &42.54& 78.49 & 76.58 &23.79&85.24&37.19 & 72.55 & 68.64\\
 \midrule

 \parbox[c]{15mm}{\multirow{2}{*}{{\parbox{2cm}{\centering Evidence-Aware}}}}
 &5 & 20  &\textbf{28.52}  & \textbf{97.16}      &\textbf{44.09} & \textbf{78.67} & \textbf{77.38}  & 24.70&86.81&38.46 & 72.93 & \textbf{69.40}\\
 &10 & 20  &28.50  & 96.82      &44.04 & 78.26 & 76.78   & 24.76&86.83&38.53 & 72.70 & 68.46\\

\bottomrule
\end{tabular}
}
 \caption{Results of the (i) sentence retrieval task in terms of Precision (P), Recall (R), and~\Fone~scores and (ii) claim verification task in terms of the label accuracy (LA) and the FEVER score evaluation metrics in dev and test sets. The ``\# Negative Examples'' column indicates the number of negative evidences that are randomly sampled for each positive instance (\ie evidence) in the training phase. The ``\# Max Instances'' column indicates the maximum number of instances that we keep for each claim (\eg evidences in the case of the pointwise loss and positive-negative pairs in the case of the pairwise loss) in the training phase. The~\cmark~symbol denotes that we use all the examples of the category (\ie ``\# Negative Examples'' or ``\# Max Instances''). The best performing models per column are highlighted in bold font. Note that the sentence retrieval metrics for both dev / test sets are reported on the 5 highly ranked evidence sentences (\ie @5).}
\label{tab:results:experiments_sentence_retrieval}
 \end{table*}

\subsection{Results}
\tabref{tab:results:experiments_sentence_retrieval} presents our results on the sentence retrieval and claim verification tasks using the various examined loss functions. 
\nikos{Regarding} the number of maximum instances, we keep as many as possible from the positive samples (\ie if we have 5 positive samples and the number of maximum instances is 20 we keep all of them, while if the number of positive samples is 25 we keep 20 of them), and then we randomly sample from the negative instances.  \revisedgiannis{The settings (\cmark,~\cmark) for the pairwise loss and settings (5,~\cmark) for the pointwise loss reported in~\tabref{tab:results:experiments_sentence_retrieval}, are similar to the ones reported in~\tabref{tab:pointwise_pairwise}}.

The evidence-aware model (see the setting with 5 negative examples and 20 maximum instances denoted as (5, 20)) is the best performing one both in \nikos{the} dev and test set in terms of FEVER score. The pairwise loss performs best in terms of label accuracy on the test set. However, the most important evaluation metric is the FEVER score, since it takes into account both the label accuracy and the predicted evidence sentences. The pointwise loss is the worst performing one when using all the evidence sentences. This is because \nikos{when} we use all the potential evidences, the number of negative samples is too large and we have a highly imbalance problem leading to low recall and FEVER score in both the dev and test set.
Note that the evidence-aware model relies on the pointwise loss (\ie the worst performing one). However, a benefit of the evidence-aware model (0.7\% in terms of FEVER score) is reported (see pointwise (5, 20)).
This showcases the important effect of ranking potential evidences simultaneously using self-attention.  
From the distance-based loss functions (\eg triplet) except for the pairwise, we observe that angular and cosine losses have worse performance compared to pairwise and triplet losses when using all the instances. This is because norm-based distance measures fit best for scoring pairs using the BERT-based representations.\\

The evidence-aware model is the best performing one (5, 20), while using only a small fraction of the overall training instances. This is because the evidence-aware model is able to take into account all possible combinations of the sampled evidences while computing attention weights. However, the same model in the (10, 20) setting showcases a reduced performance. This is due to the fact that the pointwise loss \nikos{influences} the model in a similar way as in the pointwise setting, leading to a lower performance (due to class imbalance). For the pairwise loss, we observe that the performance of the model when sampling constrained evidence sentences (see (5, 20), (10, 20) settings) is similar to the performance of the model when we do not sample evidence sentences. In addition, it seems that when one constrains the number of negative samples should also constrain the overall number of instances in order to achieve the same performance as in the non-sampling setting.
\revisedgiannis{We hypothesize that this is due to the fact that when we have a limited number of instances, it is better to have a more balanced version of the dataset.} Therefore, we conclude that the evidence-aware model achieves high performance by using few examples, and it can be \nikos{therefore} used even in the case that we have a small amount of training instances. \revisedgiannis{In the case of the pairwise loss, it is important to sample instances, otherwise it becomes computationally intensive when we take all the possible combinations between the positive and negative training instances into account.} Finally, note that it is crucial to sample negative sentences to control:
\begin{enumerate*}[label=(\roman*)]
\item the computational complexity in the case of the distance-based loss functions,
\item the memory constraints in the case of the evidence-aware model and
\item the imbalance issue in the case of the pointwise loss. 
\end{enumerate*}

\section{Challenges and Future Research Directions}

\noindent\textbf{Bias in the FEVER dataset:} As indicated in the work of~\citet{schuster:19}, there is a bias issue in the FEVER dataset that may affect the performance of the fact checking systems. Specifically, they observed that by only using the claim statement without the evidence sentences, the performance of their BERT-based system was slightly worse (8 percentage points) compared to that of an NSNM system (see~\secref{sec:methods}), which uses also the predicted evidence sentences as input. Although, these 8 percentage points difference between the two systems might be seen for some cases as a substantial improvement, this result indicates that an important part of the input is neglected. To alleviate this issue,~\citet{schuster:19} created a new test dataset, the so-called ``SYMMETRIC TEST SET''. \revisedgiannis{In this \nikos{dataset, for each claim-evidence pair, they create} a claim-evidence pair that contradicts the original one. By taking all the possible combinations (of claims and evidences), the authors of~\citet{schuster:19} were able to generate four claim-evidence pairs. \nikos{In the same work, new regularization methods to improve the performance on the new test set were introduced; for more details, we refer to~\cite{schuster:19}.}} \giannisrtwo{In a follow-up work,~\citet{schuster2021get} identified also that evidences in Wikipedia are modified over time and that the fact verification models should be able to perform adequately well when this happens. This is why they create a new dataset called VITAMINC and they focus on using contrastive evidence pairs (\ie pairs that are similar language-wise but the one supports and the other contradicts a given claim) for fact verification. They have also observed that there is a high word bias between claims and supporting evidences. This is why they wanted to minimize this in order to create difficult examples and that way robustify the fact checking models.} In the same line of research \giannisrtwo{as in the work of~\citet{schuster:19}},~\citet{thorne:20} also rely on regularization methods to improve the performance on the SYMMETRIC dataset. In another work,~\citet{karimi:20} proposed the use of product of experts~\cite{hinton:02} and focal loss~\cite{lin:17} in order to mitigate the bias in NLI systems and \nikos{reported} results on the FEVER SYMMETRIC test set. Finally, the quality of the dataset is also discussed in the work of~\citet{derczynski:20}, \nikos{which proposes two metrics that ensure the quality of the annotation of the FEVER dataset.} \revisedgiannis{In the work of~\citet{Shah_Schuster_Barzilay_2020}, augmentation techniques have been proposed in order to expand the FEVER training set using sentence-based modification approaches. The method has also been validated on the ``SYMMETRIC TEST SET''. \revisedgiannis{Similar to~\citet{schuster:19}, the work of~\citet{pratapa-etal-2020-constrained} also illustrates that LMs are able to conclude about the veracity of a claim without taking the evidences into account (\ie based on the knowledge these models have while training on large corpora). This is why they construct a new anonymized dataset, where the entities are masked out and the evidences are the only facts that are available to the model.}}\\

\noindent\revisedgiannis{\textbf{Adversarial instances:} Generating adversarial instances has been proposed in order to identify and improve the vulnerabilities of the various systems. As also described in~\secref{sec:related_tasks}, a shared-task has been organized regarding that matter~\cite{thorne:19b} and a baseline system along with scoring metrics for evaluating adversarial attacks can be found in~\citet{thorne2019adversarial,thorne:19}. Several \giannisrtwo{studies} have been proposed for introducing adversarial attacks in various datasets (see~\eg~\citet{atanasova-etal-2020-generating-label,hidey:20}), showcasing that generating adversarial instances is an interesting research direction that can help robustifying the fact checking systems.}\\

\noindent\textbf{Artificially generated dataset:} Another issue with the FEVER dataset is that the claims are artificially constructed by Wikipedia content, a source which the documents are also coming from. This way, we have a controlled environment that is far from realistic fact checking scenarios. For instance, in a realistic situation, we might assess the validity of claims coming from social media (\eg Twitter) or news sources. Therefore, for practitioners that are interested to build systems for fact checking applications (such as fake news), it is not advisable to test their system only on the FEVER dataset. \revisedgiannis{Although the dataset is artificially constructed, it can have an advantage in some cases in transfer learning scenarios. For instance, the FEVER dataset has been exploited in~\cite{wadden:20}, where a \giannisrtwo{(BERT-based)} model pre-trained on FEVER and finetuned on the SCIFACT dataset was shown to deliver  performance improvements in terms of label accuracy. However, there was no improvement in the case of sentence selection since the pre-trained \giannisrtwo{(BERT-based)} FEVER model has not been trained on a scientific corpus. Another example regarding the poor generalization performance of the FEVER dataset in transfer learning scenarios can be found in the work of~\citet{suntwal-etal-2019-importance}.} \giannisrtwo{A new dataset that also contains Wikipedia-based evidence sentences and real world claims (where user-queries from the BoolQ~\cite{clark-etal-2019-boolq} dataset have been regenerated as claims) has also been introduced. Note that although this dataset is based on Wikipedia, it contains claims that have not been constructed artificially.} \\

\noindent\revisedgiannis{\textbf{Multilinguality:} The FEVER dataset is available only in the English language. However, in case one is interested in identifying the veracity of a claim might need to consider multi-lingual content~\cite{nooralahzadeh2020zero}. This is, for instance, applicable in countries with more than one official languages. Thus, a multi-ligual dataset that can be used in the aforementioned scenario would be useful.}\\

\noindent\revisedgiannis{\textbf{Single-hop nature of the dataset:} It is worth mentioning once more (see also~\secref{sec:problem_definition}) that in 16.82\% of the FEVER claims, more than one evidence sentences are needed to conclude about their veracity. This indicates that models explicitly designed to solve more complex tasks, where multiple evidence sentences are needed to conclude about the veracity of the claim, have limited improvement compared to their full potential. For instance, see the model of~\citet{zhao:20} in which the overall performance is similar to the rest of the state-of-the-art models. However, in a more difficult subset the improvement of that model is almost 20\% in terms of FEVER score. Recently, fact checking datasets have been proposed that require more hops (sentences or documents) to validate the veracity of a claim, see for instance~\citet{jiang-etal-2020-hover,ostrowski2020multi}.}\\

\noindent\textbf{Too many negative instances:} Based on our analysis on the sentence retrieval task (see our analysis in~\secref{sec:experimental_study}), we observed that there are a lot of negative sentences that are coming from the document retrieval step. This leads to issues, such as an imbalance dataset (in the pointwise loss) and a high computational complexity (in the pairwise loss) when using all the negative samples. To circumvent these issues, following also previous research~\cite{liu:20,soleimani:20}, we randomly sample negative examples leading to lower computational time and improved performance. However, more sophisticated techniques than random sampling should be investigated to select examples that are more informative. So far, hard negative mining to select more difficult examples has been studied in the work of~\citet{soleimani:20}, although the benefit of that technique is limited. \\

\noindent\textbf{Complex vs simple models:} For the FEVER task, we observe that most of the recent research \giannisrtwo{studies} (\eg~\citet{liu:20}) focus on creating new complex architectures for the claim verification task. We conducted a small scale experiment (that is not present in~\tabref{tab:results:experiments_sentence_retrieval}), where we replaced our model for claim verification (recall that we rely on the method of~\citet{liu:20}) with a BERT-based classifier. We observed that the benefit for the pointwise loss was 0.2\%, a benefit of 0.1\% for the triplet loss and a drop of 1\% in the performance of the cosine loss when using the model of \citet{liu:20} instead of the BERT-classifier (in our early experiments on the dev set). This indicates that models relying on complex architectures without any extra knowledge have a limited benefit in the performance of the FEVER score. \revisedgiannis{Note that this gap can become larger in other datasets (see \eg the performance of the model of~\citet{liu:20,liu-etal-2020-adapting-open} on SCIFACT~\cite{wadden:20}).} On the other hand, the use of pre-trained BERT in the FEVER task~\cite{soleimani:20} gave a significant boost over the ESIM models. Moreover, the use of semantic role labeling~\cite{zhong:20} (which is an external source of knowledge) also improved the performance on the task, leading to performance similar to that of the BERT$_{\text{large}}$ model of~\citet{liu:20}. 
Finally, the combination of various models \revisedgiannis{along with the use of the GPT-3 pre-trained language model~\cite{gpt3} (\ie the main benefit is coming from that model)} in the work of~\citet{stammbach:20} leads to a substantial improvement.  
\revisedgiannis{Our experimental evidence shows that starting from a simple architecture and using one of the best performing setups in~\tabref{tab:results:experiments_sentence_retrieval} together with a BERT-based classifier for the claim verification task can deliver fair performance in the FEVER dataset. 
However, as noted above this simple classification architecture might not perform as well as more complex architectures for every dataset.}
It is also evident that an interesting direction is the use of models that bring additional knowledge (\eg new pre-trained models such as GPT-3~\cite{gpt3} or by identifying other beneficial tasks such as coreference resolution~\cite{ye2020coreferential}), \revisedgiannis{combinations of existing models or models that are able to perform complex reasoning operations in order to} extract different type of knowledge from the data. \revisedgiannis{Finally,  as  indicated  by  our performance gain, we motivate future researchers to work also on the sentence retrieval subtask, as the improvement in this subtask leads to similar improvements with architectures proposed for the claim verification subtask.}\\

\noindent\revisedgiannis{\textbf{Scalability:} Computational efficiency is important for fact-checking systems when used in real life situations such as verification of claims or fake news detection. Such systems should be able to process a huge volume of data and be able to respond fast and accurate to the queries of the users. Examples of such systems include architectures that scale well with data or the number of users, low complexity architectures, distributed training and distributed inference architectures~\cite{abrahamyan2021learned}.} \\

\noindent\revisedgiannis{\textbf{Interpretability/Explainability:} 
It is also crucial for fact checking systems to be interpretable (\ie design models that their predictions can be interpreted by \eg attention or visualising filters) and explainable (\ie design systems for post-hoc analysis of the predictions of the model).  
Although there are explainable models in the NLP community, there is little work regarding interpretability/explainability for fact checking (as one can observe in~\secref{explainability}) and the work for explainable models for the FEVER dataset is rather limited. Another direction is to study approaches for assessment of interpretability and explainability in the context of fact checking.}

\section{Conclusion}
In this paper, we \nikos{focused} on the FEVER task where the goal is to identify whether a sentence is supported or refuted by evidence sentences or if there is not enough info available, relying solely on Wikipedia documents. The aim of our work \nikos{has been} to summarize the research that has been done so far on the FEVER task, analyze the different approaches, compare the pros and cons of the proposed architectures and discuss the results in a comprehensive way. We also \nikos{conducted} a large experimental study on the sentence retrieval subtask and \nikos{drew diverse} conclusions useful for future research. \nikos{We} envision that this study will shed some light on the way that the various methods are approaching the problem, identify some potential issues with existing research and be a structured guide for new researchers to the field.

\begin{acks}
This work has been supported in the context of the MobiWave Innoviris Project.
\end{acks}

\bibliographystyle{ACM-Reference-Format}
\bibliography{references}

\end{document}